\documentclass[lettersize,journal]{IEEEtran}

\usepackage{graphics} 
\usepackage{graphicx}
\usepackage{amsmath} 
\usepackage{amssymb} 
\usepackage{cite}
\usepackage{amsfonts}
\usepackage{amsthm}
\usepackage{comment}
\usepackage[ruled,vlined,linesnumbered,noresetcount]{algorithm2e}

\usepackage{bm}
\usepackage{breqn}

\usepackage{footnote}
\makesavenoteenv{tabular}

\usepackage{caption}
\usepackage[subtle,tracking=normal]{savetrees} 
\newcommand\snapshotMargin{0.166\linewidth}

\usepackage{soul}
\usepackage{hyperref}
\usepackage[percent]{overpic}
\usepackage{xcolor}
\newcommand{\insertYoutubeLinkSim}{\url{https://youtu.be/Y44GK___QuY}}
\newcommand{\insertYoutubeLinkHardware}{\url{https://youtu.be/cihIFsPvy-Y}}

\DeclareMathOperator*{\minimize}{minimize}
\usepackage{multicol}
\usepackage{subcaption}

\title{\LARGE \bf
 Robust Quadruped Jumping via Deep Reinforcement Learning 
}

\author{Guillaume Bellegarda$^{*}$, Chuong Nguyen$^{*}$, Quan Nguyen%
\thanks{This work is supported by USC Viterbi School of Engineering.}
\thanks{Guillaume Bellegarda is with the BioRobotics Laboratory, Ecole Polytechnique Federale de Lausanne (EPFL). {\tt\footnotesize guillaume.bellegarda@epfl.ch}}
\thanks{
Chuong Nguyen and Quan Nguyen are with the Dynamic Robotics and Control Laboratory, Department of Aerospace and Mechanical Engineering, University of Southern California (USC).
        {\tt\footnotesize vanchuong.nguyen@usc.edu, quann@usc.edu}}%
\thanks{$^{*}$ The first two authors equally contributed to this work.}
}

\begin{document}
\bstctlcite{MyBSTcontrol}
\maketitle
\thispagestyle{empty}
\pagestyle{empty}

\begin{abstract} 
In this paper, we consider a general task of jumping varying distances and heights for a quadrupedal robot in noisy environments, such as off of uneven terrain and with variable robot dynamics parameters. To accurately jump in such conditions, we propose a framework using deep reinforcement learning that leverages and augments the complex solution of nonlinear trajectory optimization for quadrupedal jumping. While the standalone optimization limits jumping to take-off from flat ground and requires accurate assumptions of robot dynamics, our proposed approach improves the robustness to allow jumping off of significantly uneven terrain with variable robot dynamical parameters and environmental conditions. Compared with walking and running, the realization of aggressive jumping on hardware necessitates accounting for the motors' torque-speed relationship as well as the robot's total power limits. By incorporating these constraints into our learning framework, we successfully deploy our policy sim-to-real without further tuning, fully exploiting the available onboard power supply and motors. We demonstrate robustness to environment noise of foot disturbances of up to 6 cm in height, or $33 \%$ of the robot's nominal standing height, while jumping 2\textit{x} the body length in distance. 
\end{abstract}

\section{Introduction}

Legged robots have potential to accomplish many tasks that may be unsafe for humans, in which overcoming uneven terrain or high obstacles may be necessary. Towards real world deployment, recent works have shown highly dynamic and agile motions such as biped~\cite{AtlasBackflip2023} and quadruped~\cite{katz2019mini} backflips, wheel-legged biped jumping~\cite{BostonDynamicsHandle2016,klemm2019ascento}, quadruped running and obstacle jumping~\cite{park2017high}, and continuous jumping on stepping stones~\cite{continuous_jump}. Such methods have used either a simple model for real-time planning, or there is no associated publication. 

With respect to optimized jumping, our prior work optimizes over a full quadruped model to perform highly dynamic jumps  \cite{nguyen2019jumping}, \cite{chuongjump3D}.
A tethered quadruped model shows potential for energy efficient lunar jumping with flight phase pitch control through a reaction wheel~\cite{kolvenbach2019towards}. 
Other works have shown single legged~\cite{ding2017design} and/or dynamic miniature~\cite{haldane2016robotic,noh2012flea} jumping, for which more recent work shows SALTO performing prolonged jumping in non-laboratory settings~\cite{yim2019drift}. 

\begin{figure}[!tpb]
      \centering
      \includegraphics[width=\linewidth,trim={0cm 1cm 0cm 2.5cm},clip]{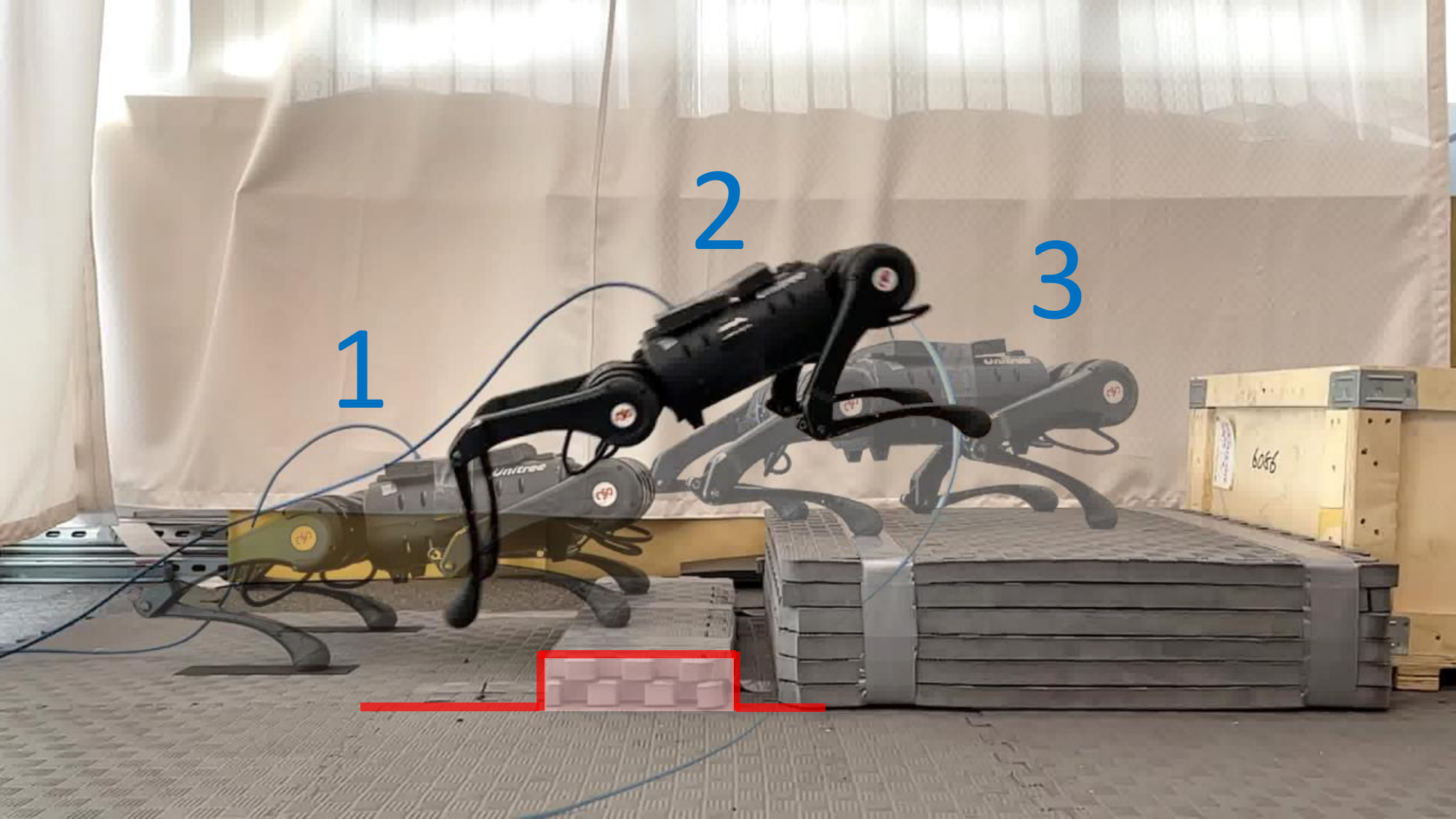} \\
      \vspace{-0.2em}
      \caption{Unitree A1, a mini quadruped robot with leg length of $0.2\ m$, successfully jumping off of an unknown $0.06\ m$ block (red line), with a goal distance of $0.6\ m$ and height $0.2\ m$.
       Experiment video link: \insertYoutubeLinkHardware.
      }
      \label{fig:intro_jump}
      \vspace{-1.5em}
\end{figure}

To make jumping more robust to external disturbances and new unseen environments, deep learning offers an attractive and generalizable formulation. 
Deep reinforcement learning in particular has recently shown impressive results in learning control policies for quadrupeds~\cite{hwangbo2019anymal, lee2020anymal,tan2018minitaur,peng2020laikagoimitation}. Typically such methods train from scratch (i.e.~use little or no prior information about the system) and rely on extensive simulation with randomized environment parameters. To facilitate the sim-to-real transfer, additional techniques are employed such as online parameter adaptation~\cite{peng2020laikagoimitation,kumar2021rma}, learned state estimation modules~\cite{ji2022concurrent}, teacher-student training~\cite{kumar2021rma,lee2020anymal,miki2022learning}, and careful Markov Decision Process choices~\cite{bellegardaIROS19TaskSpaceRL,bellegarda2021robust,bellegarda2022cpgrl}.

In this paper, we seek to use deep reinforcement learning to improve upon jumping motions produced with a trajectory optimization framework. Under ideal conditions (i.e.~starting on flat ground with a high enough coefficient of friction), the motions produced from the optimization can be accurately tracked on hardware, as shown in our prior work~\cite{nguyen2019jumping,chuongjump3D}. However, under disturbances in foot heights (i.e.~$< 0.05\ m$), a feedforward controller using only the reference trajectory will lead to taking off at an incorrect pitch angle, causing significant deviation from the desired motion. In addition to the challenges associated with highly dynamic motions, such potential errors come with high risk and can have very costly consequences such as robot damage. 

Many of the legged robots used to deploy learned locomotion policies in sim-to-real make use of direct current (DC) motors due to their ability to deliver high mobility with on-board batteries, because they offer a wide range of speeds and high torque. For learning locomotion, common assumptions include that (i) motor torque and velocity are independent in the operating region, and (ii) the on-board battery always has sufficient power to execute the learned policy (e.g. \cite{tan2018minitaur,peng2020laikagoimitation, bellegarda2021robust,bellegarda2022cpgrl}). These assumptions are reasonable for locomotion tasks such as walking or running, which require high joint velocities rather than torques, and do not need to use the full available power of the battery. In contrast to walking or running, highly aggressive motions such as jumping require both high joint torque and high joint velocity. However, DC motors typically do not allow both torque and velocity reaching their maximum values at the same time due to their inherent relationship in motor dynamics constraints. Moreover, in order to accomplish jumping motions, robots typically require total power that rapidly reaches the limits of the on-board power supply. Therefore, it is critical to consider these constraints when trying to deploy learned control policies to successfully jump in sim-to-real.  This motivates us to integrate motor dynamics and power constraints into our learning framework. Our integration considers both the torque-velocity relationship and the on-board power supply to represent the true system limits, which enables effective sim-to-real transfer for highly agile jumping motions.

\textit{Contribution:}
We present a method for improving the performance of the feedforward controller used on optimal jumping trajectories. We learn a general controller to track multiple desired jumping trajectories with deep reinforcement learning, to successfully jump in noisy environments with uneven ground, as shown in Figure~\ref{fig:intro_jump}. This controller is trained on, and able to track, many different jumping trajectories, and works with different joint gains. By learning a single controller capable of achieving many different jumps, this also avoids re-running potentially computationally expensive optimization routines at run time for relatively small differences in initial state. Moreover, in contrast to our prior work on MIT Cheetah 3~\cite{nguyen2019jumping} and Unitree A1 \cite{chuongjump3D}, our DRL controller is run as a real-time feedback controller, making our novel approach both more reliable and more robust. Importantly, we also incorporate motor dynamic constraints and power limits into the learning framework, allowing effective sim-to-real deployment without further tuning for robust and highly dynamic jumping motions. Our hardware results demonstrate that our method fully exploits the available onboard power supply and takes the motors to their limits to achieve robust and dynamic jumping motions under noisy conditions. 

The rest of this paper is organized as follows. Section~\ref{sec:background} provides background details on the robot model, reinforcement learning, and gives a brief overview of the jumping trajectory optimization. Section~\ref{sec:method} describes our learning framework design choices, including the integration of motor dynamics and power constraints to achieve robust jumping. Section~\ref{sec:result} shows extensive numerical simulations and experimental results from learning our general jumping controller, and a brief conclusion is given in Section~\ref{sec:conclusion}.

\section{Background}
\label{sec:background}

\subsection{Robot Model}
\label{sec:RobotModel}
In this paper, we validate our jumping controller on the Unitree A1~\cite{unitreeA1} quadruped robot. The A1 robot has low-inertial legs and high torque density DC motors with planetary gear reduction, and it is capable of ground force control without using any force or torque sensors. A1 uses these high-performance actuators for each hip, thigh, and knee joint to enable full 3D control of ground reaction forces. It is also equipped with contact sensors on each foot.
	
The A1 legs feature a large range of motion: the hip joints have a range of motion of $\pm$46${}^{\circ}$, the thigh joints have a range of motion from ${-60}^{\circ}$ to ${240}^{\circ}$ and the knee joints have a range from ${-154.5}^{\circ}$ to ${-52.5}^{\circ}$. Each of A1's actuators consist of a custom high torque density electric motor coupled to a single-stage 9:1 planetary gear reduction. The lower link is driven by a bar linkage which passes through the upper link. The legs are serially actuated, but to keep leg inertia low, the hip and knee actuators are co-axially located at the hip of each leg. The actuation capabilities of the A1 robot and battery power supply limits are summarized in Table \ref{tab:motor_batterry}.
\begin{table}[bt!]
	\centering
    \vspace{0.04in}
	\caption{Motor and on-board battery parameters}
	\begin{tabular}{cccc}
		\hline
		Parameter & Value & Units\\
		\hline
		Motor Gear Ratio &   9   &   \\[.5ex]
		Max Joint Torque  &  33.5 & ${Nm}$  \\[.5ex]
		Max Joint Speed    & 21  & ${Rad}/{s}$  \\[.5ex]
            Max Battery Voltage &   21.5   & ${V}$   \\[.5ex]
		Max Battery Current   &  60 & ${A}$  \\[.5ex]
            Max Battery Power & 1290 & ${W}$ \\ [.5ex]
		\hline 
		\label{tab:motor_batterry}
	\end{tabular}
    \vspace{-2em}
\end{table}

\subsection{Reinforcement Learning}
In the reinforcement learning framework~\cite{sutton1998rl}, an agent interacts with an environment modeled as a Markov Decision Process (MDP). An MDP is given by a 4-tuple $(\mathcal{S,A,P,R})$, where $\mathcal{S}$ is the set of states, $\mathcal{A}$ is the set of actions available to the agent, $\mathcal{P}: \mathcal{S} \times \mathcal{A} \times \mathcal{S} \rightarrow \mathbb{R}$ is the transition function, where $\mathcal{P}(s_{t+1} | s_t, a_t)$ gives the probability of being in state $s_t$, taking action $a_t$, and ending up in state $s_{t+1}$, and  $\mathcal{R}: \mathcal{S} \times \mathcal{A} \times \mathcal{S} \rightarrow \mathbb{R}$ is the reward function, where $\mathcal{R}(s_t,a_t,s_{t+1})$ gives the expected reward for being in state $s_t$, taking action $a_t$, and ending up in state $s_{t+1}$.
The goal of an agent is to interact with the environment by selecting actions that will maximize future rewards. In this paper, we use Soft-Actor Critic (SAC)~\cite{haarnoja2018sac} to learn the optimal policy $\pi$ to maximize jumping performance. 

SAC learns a policy, $\pi(a|s)$, and a critic, $Q_\phi(s,a)$, and aims to maximize a weighted objective of the reward and the policy entropy, $\mathbb{E}_{s_t,a_t\sim\pi} \left[\sum_t r_t + \alpha \mathcal{H}(\pi(\cdot|s_t))\right]$.  The critic parameters are learned by minimizing the squared Bellman error using transitions, $\tau_t=(s_t,a_t,s_{t+1},r_t)$, replayed from an experience buffer, $\mathcal{D}$:
\begin{equation}
    \label{eq:sac_q}
    \mathcal{L}_Q(\phi)=\mathbb{E}_{\tau\sim\mathcal{D}}\left[\big(Q_\phi(s_t,a_t) - (r_t + \gamma V(s_{t+1}))\big)^2\right]
\end{equation}
The target value of the next state can be estimated by sampling an action using the current policy:
\begin{equation}
    V(s_{t+1})=\mathbb{E}_{a'\sim\pi}\left[Q_{\tilde{\phi}}(s_{t+1},a')-\alpha \log\pi(a'|s_{t+1})\right]
\end{equation}
where $Q_{\tilde{\phi}}$ represents a more slowly updated copy of the critic.  The policy is learned by minimizing the divergence from the exponential of the soft-Q function at the same states:
\begin{equation}
    \label{eq:sac_pi}
    \mathcal{L}_\pi(\psi)=-\mathbb{E}_{a\sim\pi}\left[Q_\phi(s_t,a)-\alpha \log\pi(a|s_t)\right]
\end{equation}
This is done via the reparameterization trick for the newly sampled action, and $\alpha$ is learned against a target entropy.

\begin{figure}[!tpb]
      \centering
      \includegraphics[width=3.2in,trim={0cm 0cm 0cm 0.5cm},clip]{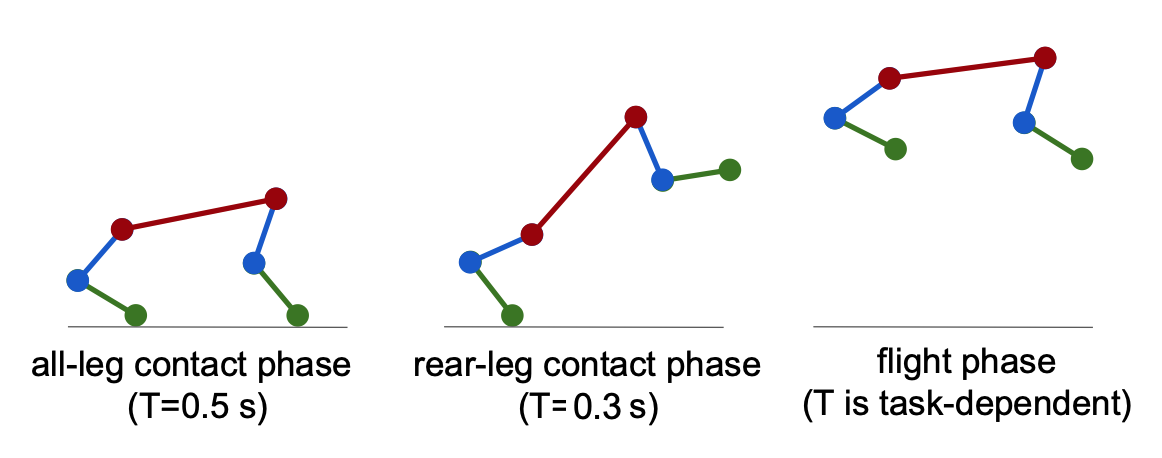} \\
      \vspace{-0.4em}
      \caption{Jumping motion phases from the trajectory optimization. }
      \label{fig:jumping_phases}
      \vspace{-1em}
\end{figure}

\subsection{Jumping Trajectory Optimization} \label{sec:learning_TO}
In this section we briefly describe the trajectory optimization framework to generate quadruped jumping motions, as well as the associated jumping controller. For full details, please see our prior work~\cite{nguyen2019jumping}. 

The robot model used in the trajectory optimization framework is a simplified sagittal plane quadruped model consisting of 5 links. The body link coordinates are represented by the position of the center of mass $[p_x, p_z]$ and the rotational angle (pitch) of the body $\theta$, while the configuration of the other links (limbs) is denoted by $\bm{q}$.  The optimization problem is divided into 3 contact phases: double contact (pre-flight preparation), single contact (rear-leg), and flight, as shown in Figure~\ref{fig:jumping_phases}. The duration of each phase is manually determined based on desired jumping distance and height. 

\begin{figure*}[!t]
      \centering
      \includegraphics[width=\linewidth]{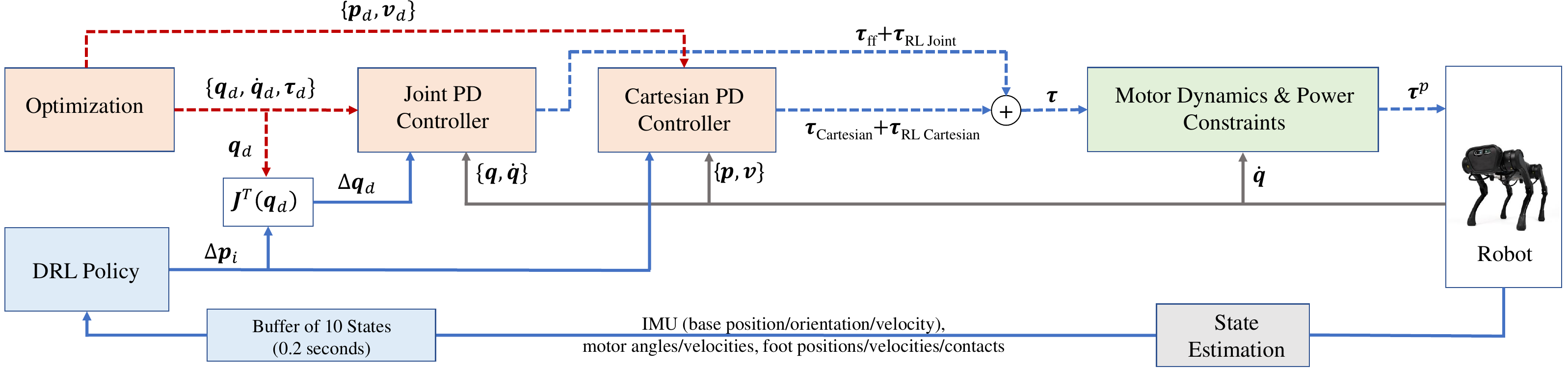} \\
      \caption{Control diagram for learning to jump robustly with reinforcement learning by leveraging the optimized jumping controller, motor dynamics, and power constraints. The combination of trajectory reference from the optimization (dotted red lines) and trajectory offsets from the DRL policy are then tracked by joint PD and Cartesian PD controllers. The dotted lines and gray lines execute at $1\ kHz$, while the solid blue lines execute at $50\ Hz$ during contact phases. The motor dynamics and power constraints module is designed to represent the hardware capability, which limits the final torque reference $\bm{\tau}^{p}$ for the robot's motors.} 
      \label{fig:control_diagram}
      \vspace{-1em}
\end{figure*}

At a high level, the resulting discrete time optimization can be formulated as follows: 
\begin{align*}
\mbox{$\displaystyle{ \minimize_{\bm{x}_k,\bm{u}_k;  \ k=1...N } }$ } \ & J\bigl(\bm{x}_N\bigr) + h\sum_{k=1}^{N} w\bigl(\bm{x}_k,\bm{u}_k\bigr) \\
\mbox{subject to} \ & d\bigl(\bm{x}_k,\bm{u}_k,\bm{x}_{k+1}) = 0, \ k=1...N-1 \\
& \phi\bigl(\bm{x}_k,\bm{u}_k\bigr) = 0, \ k=1...N \\ 
& \psi\bigl(\bm{x}_k,\bm{u}_k\bigr) \leq 0, \ k=1...N 
\end{align*}
where $\bm{x}_k=\left[p_{x,k};~p_{z,k};~\theta_{k};~\bm{q}_{k}\right]$ is the full state of the system at sample $k$ along the trajectory, $\bm{u}_k$ is the corresponding control input, $J$ and $w$ are final and additive costs to jump to a particular height and distance while minimizing energy, $h$ is the time between sample points $k$ and $k+1$, and $N$ is the total number of samples along the trajectory. The constraints are specified as follows:
\begin{itemize}
    \item The function $d(\cdot)$ captures the full-body dynamic constraints ~\cite{nguyen2019jumping}, which is discretized from
    \begin{equation}\label{eq:full_dynamics_2D}
    \begin{bmatrix}\bm{M} & -\bm{J}_c^T \\ -\bm{J}_c^T & \mathbf{0} \end{bmatrix} \begin{bmatrix}
    \bm{\ddot{x}} \\ \bm{f}_c
    \end{bmatrix}= \begin{bmatrix}-\bm{C}\bm{\dot x}-\bm{g} + \bm{S}\bm{\tau}+\bm{S}_{f}\bm{\tau}_{f} \\ \bm{\dot J}_{c}(\bm{x})\bm{\dot x}\end{bmatrix}, \nonumber
    \end{equation}
    where $\bm{M}$ is the mass matrix, $\bm{C}$ represents the Coriolis  and centrifugal terms, $\bm{g}$ denotes the gravity vector, $\bm{J}_c$ is the spatial Jacobian expressed at the foot contact, $\bm{S}$ and $\bm{S}_{fric}$ are distribution matrices of actuator torques $\bm{\tau}$ and joint friction torques $\bm{\tau}_{fric}$, $\bm{f}_c$ is the spatial force at the foot contact. The dimensions of $\bm{J}_c$ and $\bm{f}_c$ depend on the contact phases.
    
    \item The function $\phi(\cdot)$ represents equality constraints on initial joint and body configurations, pre-landing configuration, and final body configuration. 
    
    \item The function $\psi(\cdot)$ captures inequality constraints including joint angle/velocity/torque limits, friction cone limits, minimum ground reaction forces, and geometric constraints related to the ground and obstacle clearance. 
\end{itemize}
    
The optimization produces desired joint angles $(\bm{q}_d)$, joint velocities $(\bm{\dot{q}}_d)$ and feed-forward joint torques $(\bm{\tau}_d)$ at a sampling time of 10 \emph{ms}, which are then linearly interpolated to 1 \emph{ms}. These can be tracked by the following joint PD controller running at 1 \emph{kHz} as:
\begin{align}
    \bm{\tau}_{\mathrm{ff}} = \bm{K}_{p,joint} (\bm{q}_d -\bm{q}) + \bm{K}_{d,joint}  (\bm{\dot{q}}_d -\bm{\dot{q}}) + \bm{\tau}_d 
    \label{eqn:jumping_jointPD}
\end{align}
where $\bm{K}_{p,joint}$ and $\bm{K}_{d,joint}$ are diagonal matrices of proportional and derivative gains in the joint coordinates. 

To improve tracking performance, a Cartesian PD controller is added. From the desired joint angle $(\bm{q}_d)$ and joint velocity $(\bm{\dot{q}}_d)$ trajectories, we extract desired foot positions $(\bm{p}_d)$ and foot velocities $(\bm{v}_d)$ in the leg frame. Thus the Cartesian PD and full controllers for tracking the desired jumping trajectory become: 
\begin{align}
    \bm{\tau}_{\mathrm{Cartesian}} &= \bm{J}(\bm{q})^\top \left[ \bm{K}_p \left(\bm{p}_d - \bm{p} \right) + \bm{K}_d \left(\bm{v}_d -  \bm{v} \right)  \right]\\
    \bm{\tau}_{\mathrm{opt}} &= \bm{\tau}_{\mathrm{Cartesian}} + \bm{\tau}_{\mathrm{ff}}
    \label{eqn:jumping_full}
\end{align}
where $\bm{J}(\bm{q})$ is the foot Jacobian at joint configuration $\bm{q}$, $\bm{K}_p$ and $\bm{K}_d$ are diagonal matrices of proportional and derivative gains in Cartesian coordinates, and $\bm{\tau}_{\mathrm{ff}}$ is the feed-forward torque from Equation~\ref{eqn:jumping_jointPD}. 

\section{Robust Jumping with Reinforcement Learning}
\label{sec:method}

Given the already established Cartesian and joint space controller for tracking jumping motions, in this section we describe our process and reinforcement learning framework for learning to modify and track these optimal trajectories in the presence of environmental noise and disturbances. Our learning framework integrates motor dynamics and power constraints, taking into account the motor torque-speed relationship and maximum on-board power supply, to represent the actual system limits. This integration enables effective sim-to-real transfer for robust and agile jumping on the robot hardware.

\subsection{Learning Optimal Trajectory Offsets}
In order to provide an intuitive mapping between actions and their effects on the system, we propose learning to appropriately offset the jumping trajectories to cope with disturbances in the environment. Specifically, we consider learning in Cartesian space, with the idea that the agent can more directly observe the effects of its actions, as well as more easily map offsets based on the environmental observation than it can in joint space.  
In particular, we consider learning Cartesian space offsets $( \Delta \bm{p}_{RL})$ to modify the existing optimal trajectory, which will be combined with the existing jumping controller in Equation~\ref{eqn:jumping_full}. The corresponding torque contribution from these Cartesian space offsets can be written as:
\begin{align}
    \bm{\tau}_{\mathrm{RL\ Cartesian}} =  \bm{J}(\bm{q})^\top \left[ \bm{K}_p \left( \Delta \bm{p}_{RL} - \bm{p} \right)   \right] 
    \label{eqn:jumping_rl_cartesian}
\end{align}

Such Cartesian space offsets could result in significant deviations in the optimal joint trajectories $(\bm{q}_d)$. To avoid joint space gain feedback from counteracting these deviations, we also add offsets in joint space corresponding to those desired in Cartesian space as:
\begin{align}
    \Delta \bm{q}_{RL} =  \bm{J}(\bm{q}_d)^\top \Delta \bm{p}_{RL}
\end{align}
This makes the joint space reinforcement learning torque contribution as follows: 
\begin{align}
    \bm{\tau}_{\mathrm{RL\ Joint}} =  \bm{K}_{p,joint} (\Delta \bm{q}_{RL} -\bm{q})
    \label{eqn:jumping_rl_joint}
\end{align}

The full controller for tracking the desired trajectories with learned reinforcement residual offsets is then the summation of the original jumping controller (Equation~\ref{eqn:jumping_full}) with the offset contributions (Equations~\ref{eqn:jumping_rl_cartesian} and~\ref{eqn:jumping_rl_joint}):
\begin{align}
    \bm{\tau} = \bm{\tau}_{\mathrm{opt}} +  \bm{\tau}_{\mathrm{RL\ Cartesian}} + \bm{\tau}_{\mathrm{RL\ Joint}} 
    \label{eqn:jumping_rl_full}
\end{align}

\subsection{Reinforcement Learning Details}

There are several challenging aspects of jumping that motivate our observation space, action space, and reward function choices.  Firstly, while in the air, it is very difficult to meaningfully adjust body position or orientation. Secondly, even small noise or deviations in the trajectory before take-off can have a large effect on landing location and orientation. Thirdly, the entire motion happens very quickly, with the pre-jump phase taking only 0.8 $s$, and the flight phase roughly 0.4 $s$, depending on distance and height. To mitigate these issues, we learn actions to modulate movement while in contact only, and apply $\bm{\tau}_{\text{opt}}$ in the air. Figure~\ref{fig:control_diagram} shows the full block diagram for integrating our reinforcement learning action with the optimal jumping controller, and we describe the MDP below.

\textit{Action Space}:
The action space consists of the desired trajectory offsets $\mathbf{a}  \in \mathbb{R}^{12} $ for the contact phase of the jumping trajectory, which are updated at $50\ Hz$. The agent chooses offsets in $[-0.05,0.05]\ m$ from each foot's local desired $(x,y,z)$ Cartesian positions from the optimization.

\textit{Observation Space}: 
The observation space consists of the full robot state at the initial state, trajectory end state (goal), as well as a history of states in the previous 0.2 $s$. The history of states is a stack of 10 observations updated at $50\ Hz$. Each of these states consists of: body state (position, orientation, linear and angular velocities), joint state (positions, velocities), foot state (positions, velocities), and foot contact booleans. All values are first normalized before being used for training purposes by reinforcement learning. 

\textit{Reward}: 
We give a sparse, single reward at the end of the jumping trajectory based on the error between the desired and actual landing position and orientation. We give a sparse reward, rather than dense rewards for tracking the optimized jumping trajectory at every time step, as significant deviations to the offline optimized trajectory can be expected (and will be needed) for large environmental noise. 

More precisely, the reward function attempts to minimize deviations in the body position  $(x_b,y_b,z_b)$ and orientation $(\phi_b,\theta_b,\psi_b)$ from the final desired states in the optimal trajectory: body position $(x_N,y_N,z_N)$ and orientation $(\phi_N,\theta_N,\psi_N)$. The final orientation $(\phi_N,\theta_N,\psi_N)$ is always $(0,0,0)$ as we would like the agent to land upright at its standing orientation. The reward function is written as: 
\begin{align}
\label{eq:min_dist}
R(s_t,a_t,s_{t+1}) = w(1 &- \| (x_b,y_b,z_b) - (x_N, y_N, z_N) \| \nonumber \\ 
&- \| (\phi_b,\theta_b,\psi_b)) \| )
\end{align}
where $w$ is a terminal weight. This reward scheme ensures a reward of $w$ for perfect tracking, and will decrease from there, and even be negative, for very poor tracking.

\subsection{Training Details}
\label{sec:training_details}

We first generate 13 jumping trajectories, with final desired positions ranging in distance in $[0.5, 0.8]\ m$ and in height in $[0, 0.4]\ m$. At the beginning of each episode, one of the trajectories is randomly selected to track, and random noise is added to the environment. The noise consists of blocks up to $0.1\ m$ in height under each foot,  and the body mass and inertia are each varied randomly by up to 5\% of their nominal values. 

We use PyBullet~\cite{pybullet} as the physics engine for training and simulation purposes, and the A1 quadruped model introduced in Sec.~\ref{sec:RobotModel}. %
For SAC~\cite{haarnoja2018sac}, our neural networks are multi-layer perceptrons with two hidden layers of 512 neurons each, with tanh activation. Other training hyperparameters are listed in Table~\ref{table:tab1}. 

\begin{table}[tpb]
\centering
\caption{SAC Hyperparameters.}
\begin{tabular}{ c c  }
Parameter & Value \\
\hline
optimizer & Adam  \\
learning rate & $3 \cdot 10^{-4}$ \\
discount ($\gamma$) & 0.99 \\
replay buffer size & $10^6$ \\
initial steps & 1000 \\
number of hidden layers (all networks) & 2 \\
number of hidden units per layer & 512 \\
nonlinearity & tanh \\
batch size & 64 \\
target smoothing coefficient ($\tau$) & 0.005 \\
target update interval & 1 \\
gradient steps & 1 \\
\hline
\end{tabular} \\
\label{table:tab1}
\vspace{-1em}
\end{table}

\subsection{Motor Dynamics and Power Constraints}
\label{sec:MDC}
Since legged robots must rapidly reach their motor and on-board power supply limits to accomplish dynamic jumping maneuvers, it is critical to model and integrate the motor dynamics and power constraints during training to represent the true system limits.  This integration in turn limits the reference torque individually applied to each motor, which enables successful sim-to-real transfer for aggressive motions such as jumping. 

\subsubsection{Motor and Power Modeling}
First, we revisit a simplified DC motor model which captures the inherent torque-velocity relationship.
Since the inductance of stator windings is typically small (approximately $1 mH$ for an A1 robot motor \cite{unitreeA1}), the voltage applied to each motor $i \in \lbrace1, ..., n \rbrace $ can be simplified as follows
\begin{align} \label{eq:voltage_cs}
V_i(\tau_i^m, \dot{q}_i^m)=I_i^m(\tau_i^m) R_i+\varphi_i(\dot{q}_i^m),
\end{align} 
where $R_i$ is the resistance of the coil windings, and $\dot{q}_i^m$ is the motor velocity. The  back electromotive force (EMF) of the windings generated by the rotation of the motor is estimated by $\varphi_i(\dot{q}_i^m) = K_v\dot{q}_i^m$, and the current $I_i^m(\tau_i^m)$ flowing in the windings relates to the motor torque via $I_i^m = \tau_i^m /K_{\tau} $. Here, $K_v$ and $K_{\tau}$ are the electric motor velocity constant and torque constant, respectively. Considering the gear ratio $g_r$ which relates
\begin{equation}
    \tau_i = \tau_i^m g_r, ~ \dot{q}_i = \dot{q}_i^m/ g_r \nonumber
\end{equation}
we can rewrite the voltage equation (\ref{eq:voltage_cs}) as a linear combination of joint torque and joint velocity as
\begin{align} \label{eq:}
V_i(\tau_i, \dot{q}_i)=\alpha \tau_i+ \beta \dot{q}_i,
\end{align} 
where $\alpha=R_i/(K_\tau g_r)$ and $\beta=K_vg_r$, respectively.

Moreover, it is noteworthy that jumping maneuvers are highly demanding and normally quickly drain the battery's power capacity. Hence, it is also essential to consider the total power required to run all of the robot's motors in the learning framework. The total power can be estimated by:
\begin{align} \label{eq:MDC_power}
    \sum_{i=1}^n P_i(\tau_i, \dot{q}_i) = \sum_{i=1}^n V_i I_i= \sum_{i=1}^n \left( \frac{\tau_i}{K_{\tau} g_r} \right)^2 R_i + \frac{K_{v} \tau_i \dot{q}_i}{K_{\tau} }
\end{align}
The power consists of two parts: the first part is power dissipation on the windings, which is proportional to $\tau_i^2 R_i$; the second part is the power of rotation $\tau_i \dot{q}_i$. 

Having revisited the torque-speed relationship and total power estimation, we will propose and integrate motor dynamic constraints and power limits into the simulation environment.
\subsubsection{Implementation of Motor Dynamics and Power Constraints}
We incorporate the motor dynamics and power constraints into the DRL framework in order to enforce restrictions on the final reference torque that is applied to the motors, as depicted in Fig. \ref{fig:control_diagram}. In particular, the final reference needs to satisfy the following conditions:
\begin{itemize}
    \item [i.] \textit{Motor dynamic constraints (MDC)} establish a key relationship between joint torque and velocity in conjunction with the available voltage supply capability  $V_{bat}$, i.e.,
    \begin{align} \label{eq:MDC_voltage}
        | V_i(\tau_i, \dot{q}_i)|=\lvert \alpha \tau_i+ \beta \dot{q}_i \rvert \leq V_{bat}
    \end{align}
    \item [ii.] \textit{Power limits}: The total power supplied to all $n$ motors is constrained by the power supply capability. This requires that the total power for operating all motors is limited by the battery power $P_{bat}$, i.e.,
    \begin{align} \label{eq:limits_power}
        \sum_{i=1}^n P_i(\tau_i, \dot{q}_i) \leq P_{bat}
    \end{align}
\end{itemize}

It is noted that the motor dynamic constraints (\ref{eq:MDC_voltage}) imply that the joint torques and joint velocities  cannot simultaneously reach their respective limits. In particular, the DC motor reaches maximum velocity when running at no load, and the back EMF approaches the supply voltage. Approximately, $\dot{q}_i^{max}=V_{bat}/\beta$, giving rise to the following constraints: 
\begin{align} \label{eq:max_vel}
    V_{bat} \geq  \beta\dot{q}_i, ~ -V_{bat} \leq  \beta\dot{q}_i
\end{align}
With these conditions established, we integrate the motor dynamics and power constraints  into the simulation environment, as described in Algorithm \ref{alg:mdc_powerlimits}.

\begin{algorithm}
\caption{Integration of Motor Dynamics and Power Constraints into the Simulation Environment}
\label{alg:mdc_powerlimits}
\textbf{Input}: The total torque $\bm{\tau}$ (Eq. \ref{eqn:jumping_rl_full}), feedback joint velocity $\bm{\dot{q}}$, battery and motor parameters.\\
\textbf{Output}: Final reference torque for each motor \\
\textit{~~~~~~~~~~~~~~~(i) Motor Dynamic Constraints (MDC):}
\raggedright
\SetAlgoLined
\LinesNumbered
\SetKwProg{Function}{function}{}{end}
\SetKwRepeat{Do}{do}{until}

Compute: $V_i \gets \alpha \tau_i+ \beta \dot{q}_i $\
\textbf{if} $V_i> V_{bat}$ \textbf{then} $\tau_i^v \gets \left (V_{bat}-\beta \dot{q}_i \right)/\alpha $ ~~~~~~~~~~~ \\
\textbf{else if} $V_i< -V_{bat}$ \textbf{then} $\tau_i^v \gets \left (-V_{bat}-\beta \dot{q}_i \right)/\alpha$ \\
\textbf{else} $\tau_i^v =\tau_i$ ~~~~~~~~~~~~~~~~~~~~~~~~~~~~~~~~~~~~~~~~~~~~~~\\
\Return $\bm{\tau}^v$\; 

\textit{~~~~~~~~~~~~~~~~~~~(ii) Power Limits:}\\
\raggedright
\SetAlgoLined
\LinesNumbered
\SetKwProg{Function}{function}{}{end}
\SetKwRepeat{Do}{do}{until}
Compute $P_{total}^d \gets \sum_{i=1}^n P_i(\tau_i^v, \dot{q}_i)$\\
\uIf{$P_{total}^d > P_{bat}$}{%
  $\tau_i^{p} \gets \eta \tau_i^v $: reduce torque proportionally ($0<\eta<1$)\\ 
   where $ \eta = \frac{\sqrt{B^2+4AP_{max}}-B}{2A}$, \\
}\ElseIf{$P_{total}^d \leq P_{bat}$}{
  $\tau_i^{p} \gets  \tau_i^v$\;
}
\Return $\bm{\tau}^p$\;
\end{algorithm}

Algorithm \ref{alg:mdc_powerlimits} should be executed sequentially, starting with the MDC block, followed by the power limits. In case the torque $\tau_i^v$ obtained from the MDC block violates the power constraints (\ref{eq:limits_power}), we will proportionally decrease this torque by setting $\tau_i^{p}= \eta \tau_i^v$. This modified reference will then be utilized for the motor in simulation. Consequently, the task is to find a value of $\eta \in (0,1)$ that satisfies the following quadratic equation:
\begin{align} \label{eq:MDC_power2}
      f(\eta) \triangleq \eta^2 \sum_{i=1}^n  \frac{R_i (\tau_i^v)^2}{K_{\tau}^2 g_r^2} + \eta \sum_{i=1}^n \frac{\tau_i^v \dot{q}_i}{K_{\tau} K_{v}} = P_{bat}
\end{align}
Let $A=\sum_{i=1}^n  \frac{R_i (\tau_i^v)^2}{K_{\tau}^2 g_r^2}$ and $B=\sum_{i=1}^n \frac{\tau_i^v \dot{q}_i}{K_{\tau} K_{v}}$, then (\ref{eq:MDC_power2}) yields:
\begin{align} \label{eq:MDC_power3}
    \eta = \frac{\sqrt{B^2+4AP_{max}}-B}{2A}
\end{align}

In the following, we will prove that the final output $\bm{\tau}^p$ at the end of Algorithm 1 also satisfies the voltage constraints in Equation (\ref{eq:MDC_voltage}), i.e., $|V_i(\tau_i^p,\dot{q}_i)| \leq V_{bat}$, $\forall i \in \lbrace 1,...,n \rbrace$. 
Indeed, if $P_{total}^d \leq P_{bat}$, then Algorithm 1 assigns $\tau_i^{p} =\tau_i^v$. This output torque $\tau_i^{p}$ trivially satisfies the voltage constraints
\begin{align} 
    | V_i(\tau_i^p, \dot{q}_i) | = \lvert \alpha \tau_i^p+ \beta \dot{q}_i \rvert = \lvert \alpha \tau_i^v+ \beta \dot{q}_i \rvert \leq V_{bat}
\end{align}
Therefore, it remains to be shown that $|V_i(\tau_i^p,\dot{q}_i)| \leq V_{bat}$ when $\tau_i^{p}=\eta \tau_i^v, \ \eta \in (0, 1)$ for the case $P_{total}^d > P_{bat}$.

We start with the voltage values, which are obtained from the MDC block: $| V_i(\tau_i^v, \dot{q}_i)|=\lvert \alpha \tau_i^v+ \beta \dot{q}_i \rvert \leq V_{bat}$. This inequality is equivalent to $\left (-V_{bat}-\beta \dot{q}_i \right) /\alpha \leq \tau_i^v \leq \left (V_{bat}-\beta \dot{q}_i \right)/ \alpha$. Then, by multiplying all sides of the inequalities by $ \alpha \eta$, we obtain $\eta \left (-V_{bat}-\beta \dot{q}_i \right) \leq \alpha \tau_i^p \leq  \eta \left (V_{bat}-\beta \dot{q}_i \right)$. Therefore, adding the term $\beta \dot{q}_i$ to the inequalities yields
\begin{align}
    (1-\eta)\beta \dot{q}_i-\eta V_{bat} \leq V_i^p \leq (1-\eta)\beta \dot{q}_i+\eta V_{bat} 
\end{align}
    
Combining with Eq. (\ref{eq:max_vel}) and $0 < \eta <1$, one can verify that
    \begin{subequations}
    \begin{align}
       V_i^p -V_{bat} &\leq (1-\eta)\left(\beta \dot{q}_i-V_{bat}  \right) \leq 0\\
       V_i^p +V_{bat} &\geq (1-\eta)\left(\beta \dot{q}_i+V_{bat}  \right) \geq 0
    \end{align}
    \end{subequations}
This yields $|V_i^p| \leq V_{bat}$, $\forall i \in \lbrace 1,...,n \rbrace$, and the proof is complete.

In conclusion, our proposed algorithm computes final torque $\bm{\tau}^p$ that theoretically guarantees both motor dynamic constraints (Eq. \ref{eq:MDC_voltage}) and power limits (Eq. \ref{eq:limits_power}). In hardware experiments, we will verify the key role of these constraints in attaining effective sim-to-real transfers.

\section{Results}
\label{sec:result}
\begin{figure*}[th]
      \centering
      \includegraphics[width=\snapshotMargin]{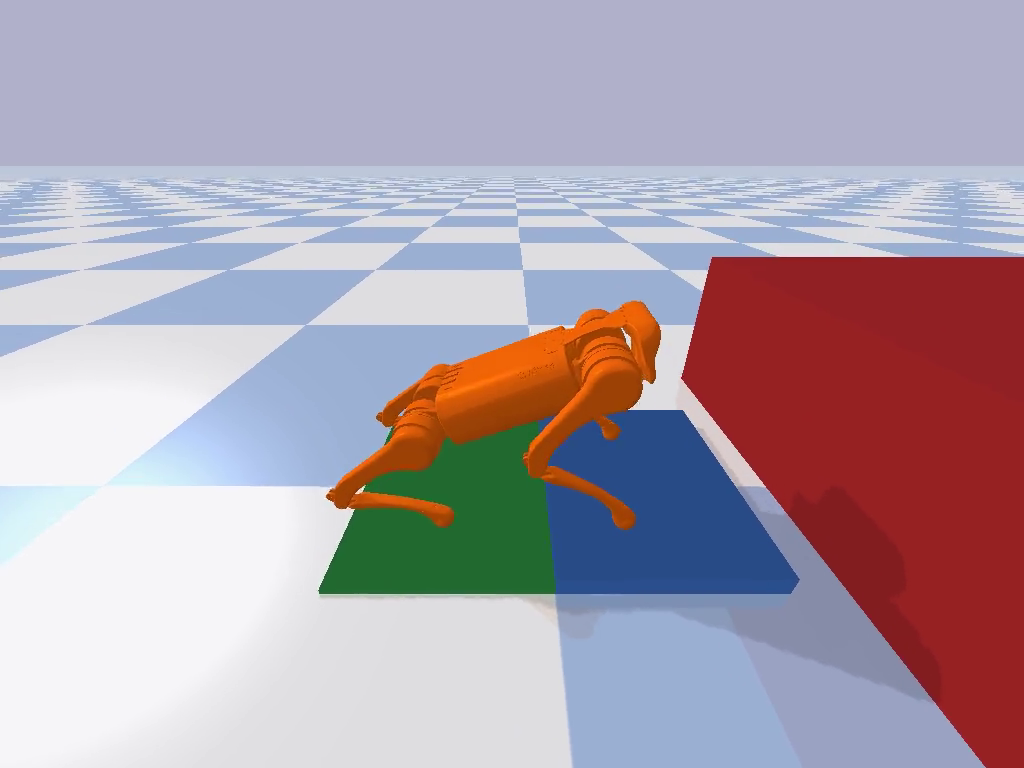}\includegraphics[width=\snapshotMargin]{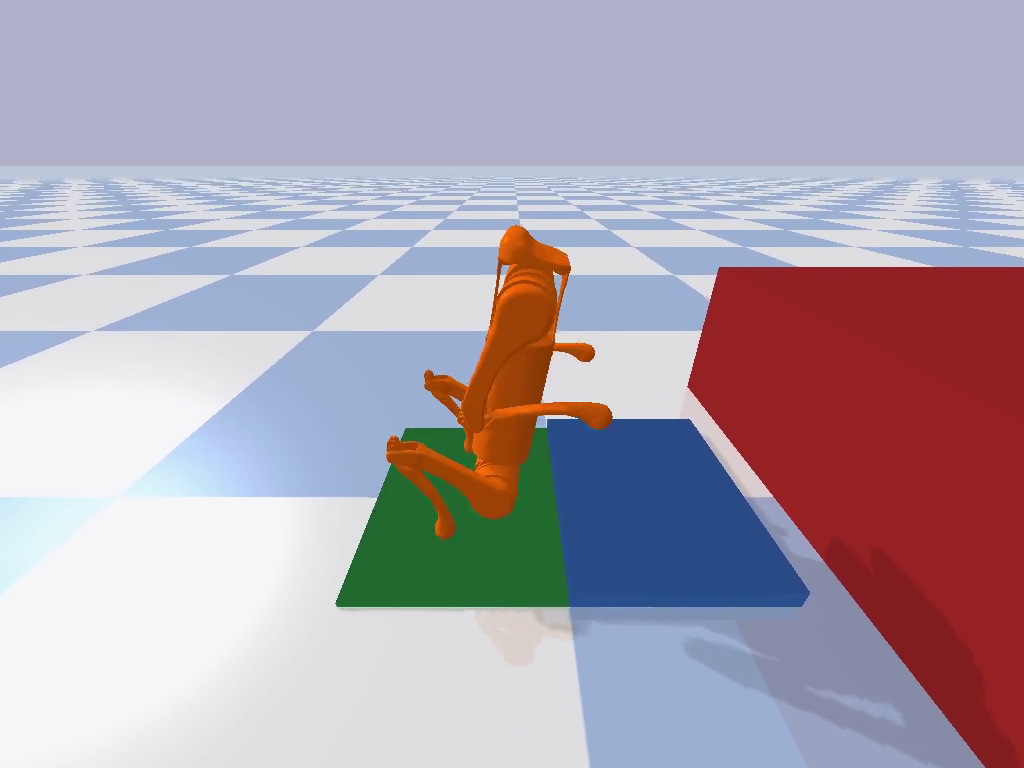}\includegraphics[width=\snapshotMargin]{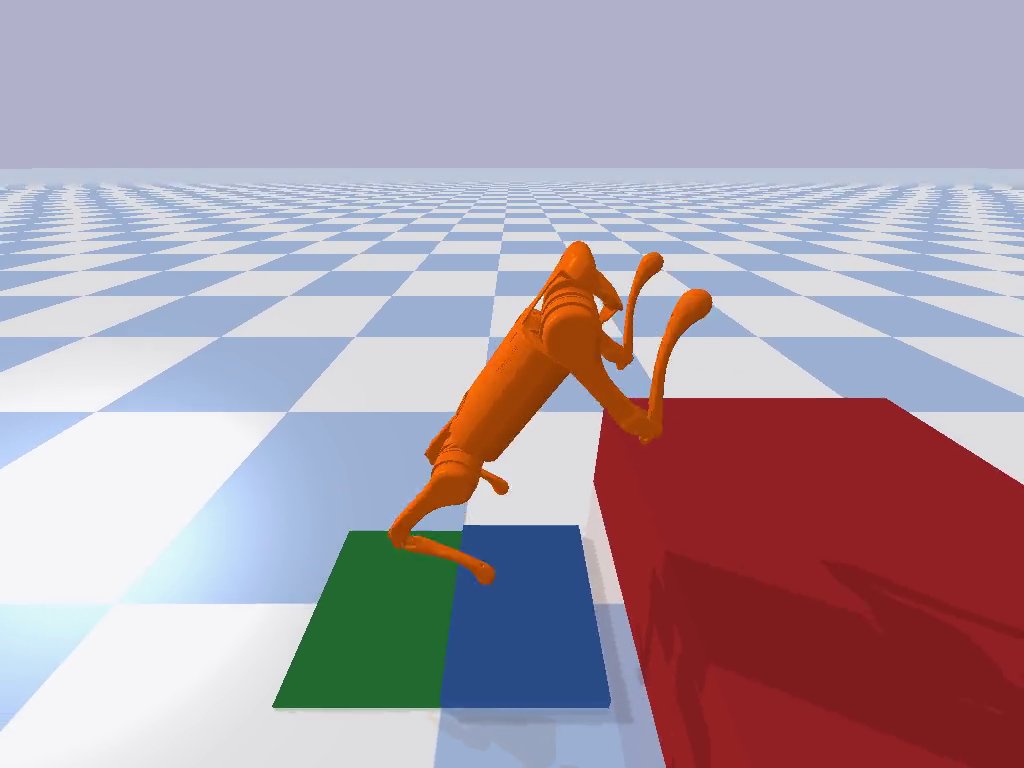}\includegraphics[width=\snapshotMargin]{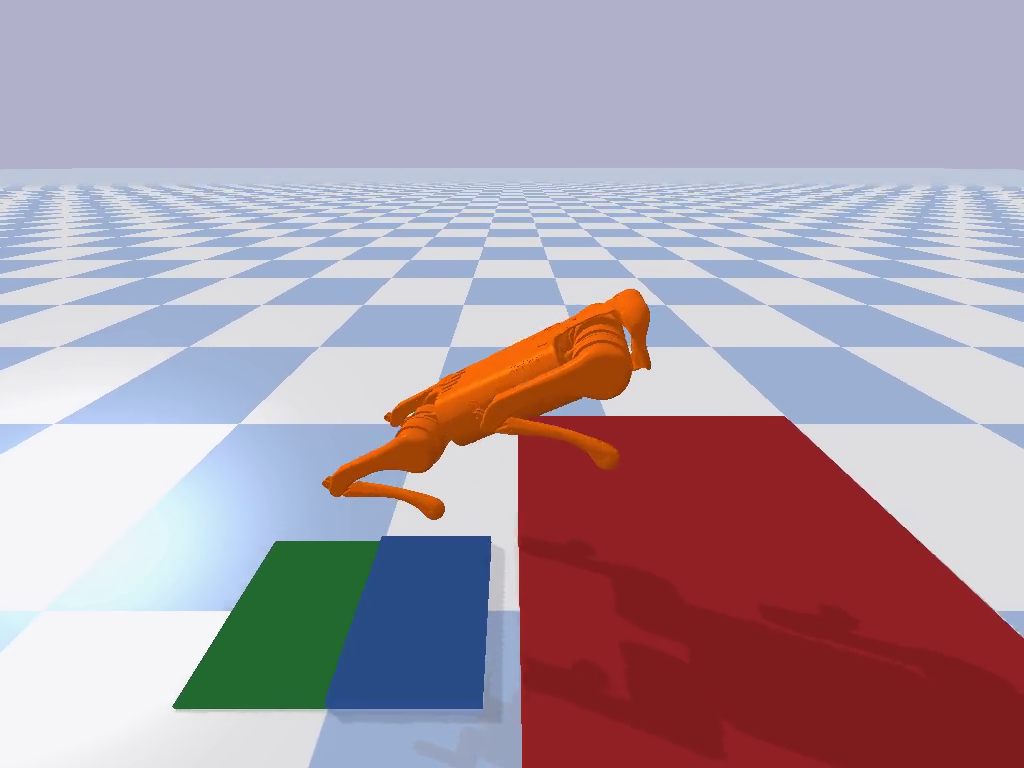}\includegraphics[width=\snapshotMargin]{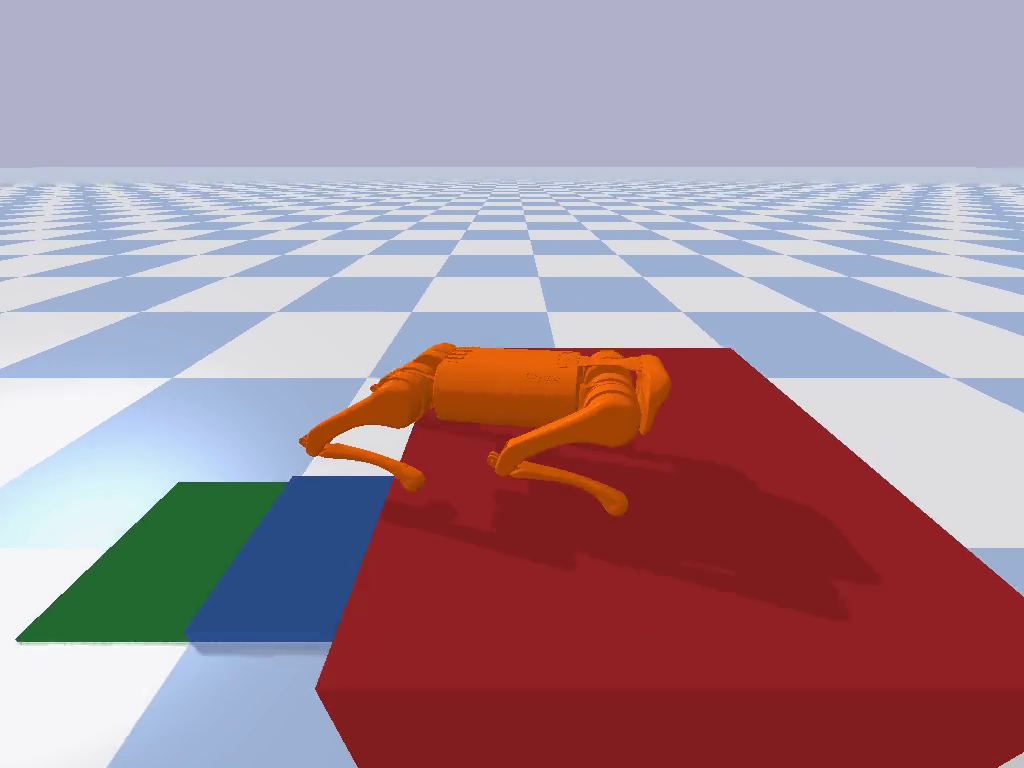}\includegraphics[width=\snapshotMargin]{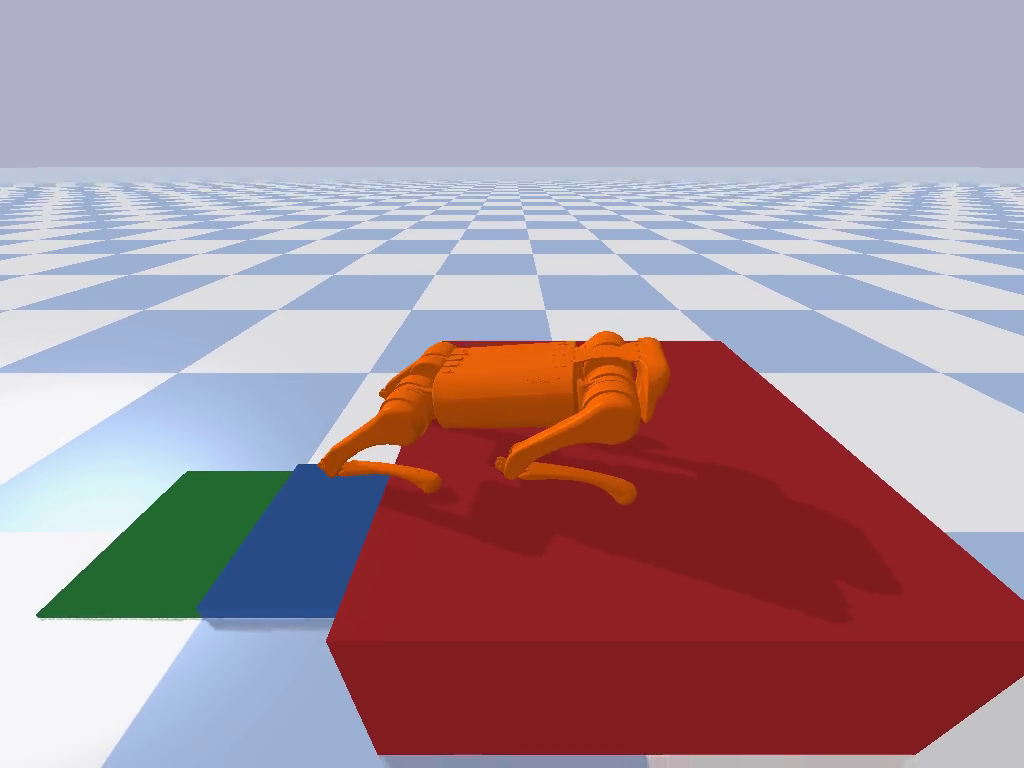}\\
      \vspace{0.1em}
      \includegraphics[width=\snapshotMargin]{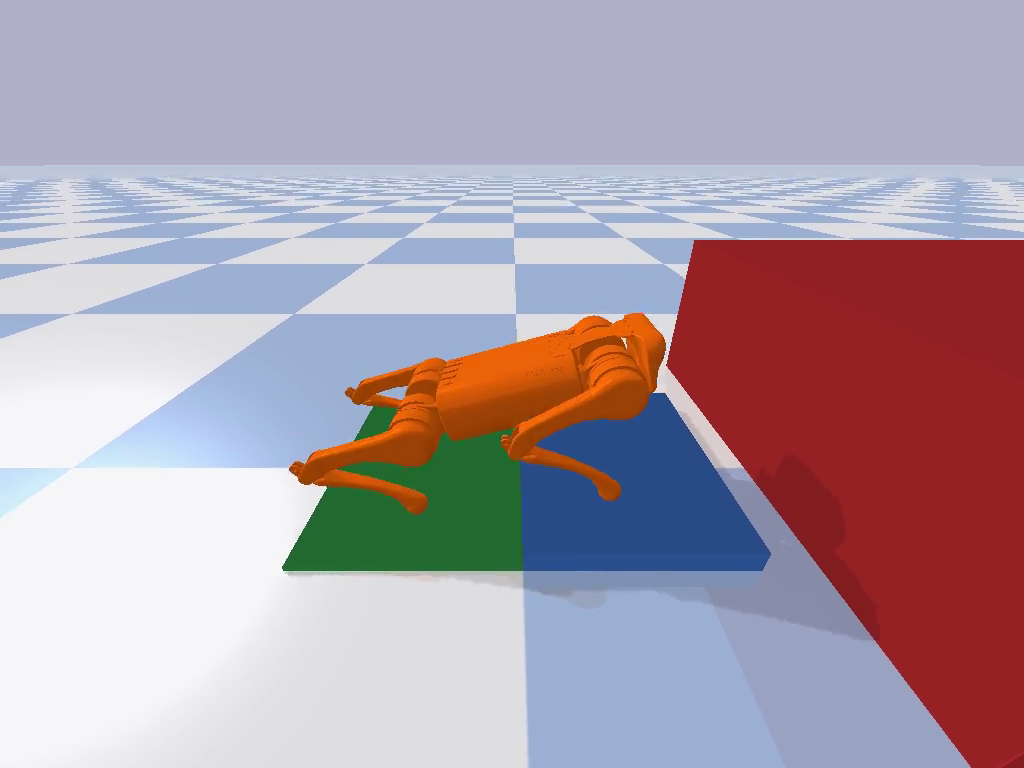}\includegraphics[width=\snapshotMargin]{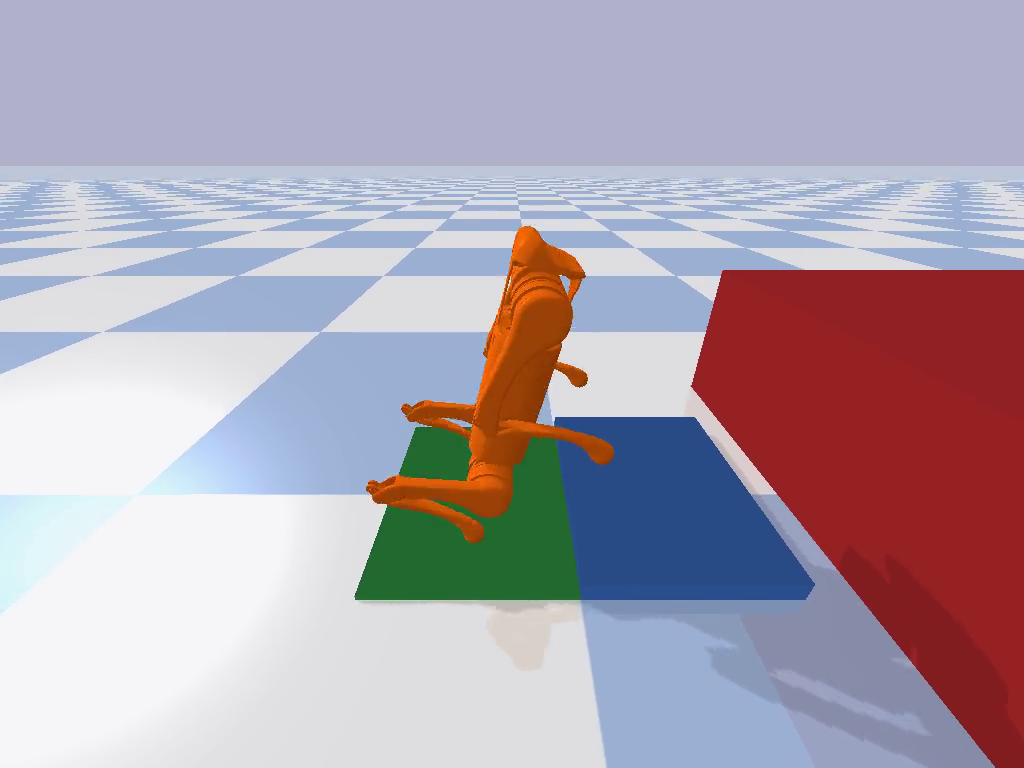}\includegraphics[width=\snapshotMargin]{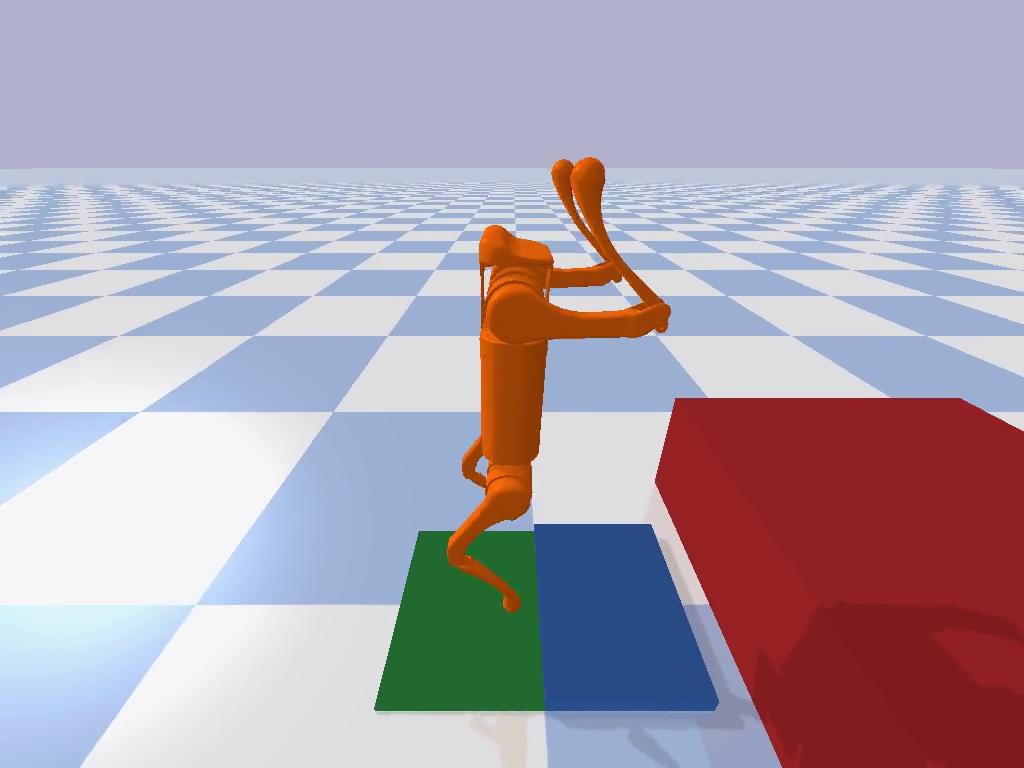}\includegraphics[width=\snapshotMargin]{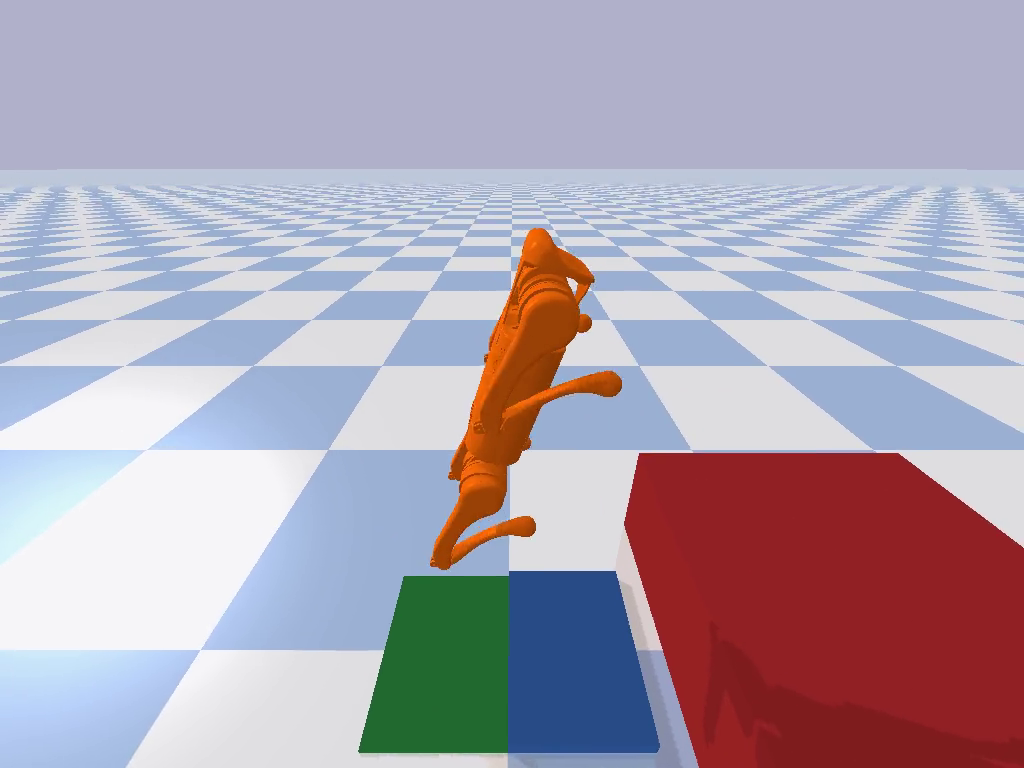}\includegraphics[width=\snapshotMargin]{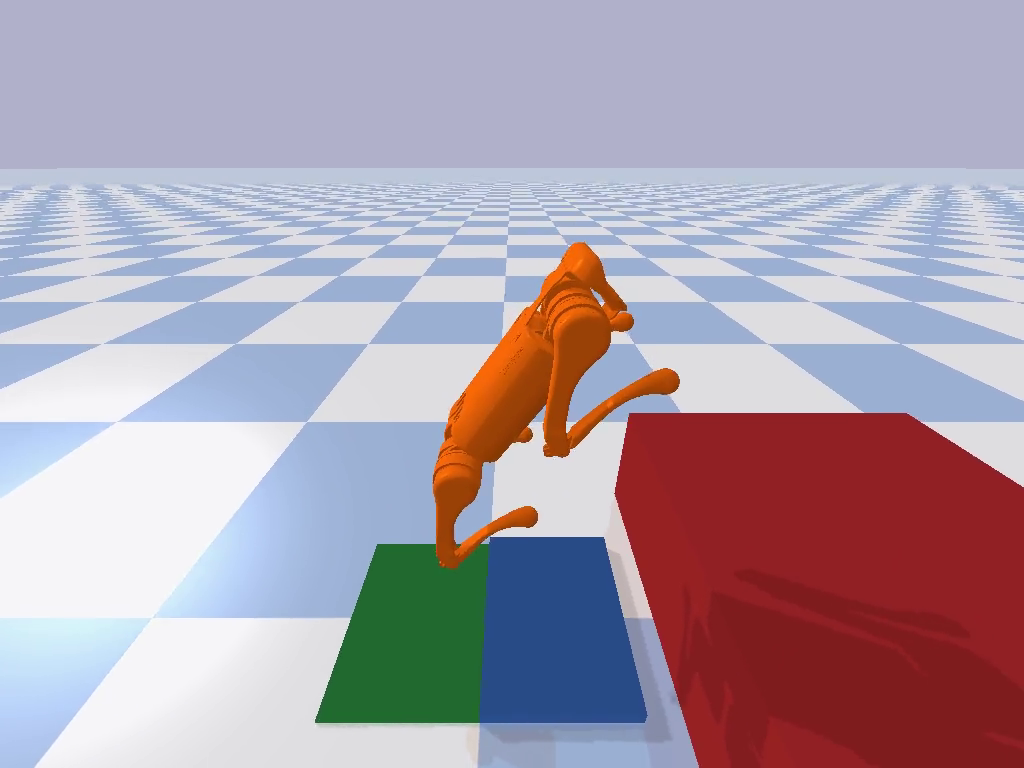}\includegraphics[width=\snapshotMargin]{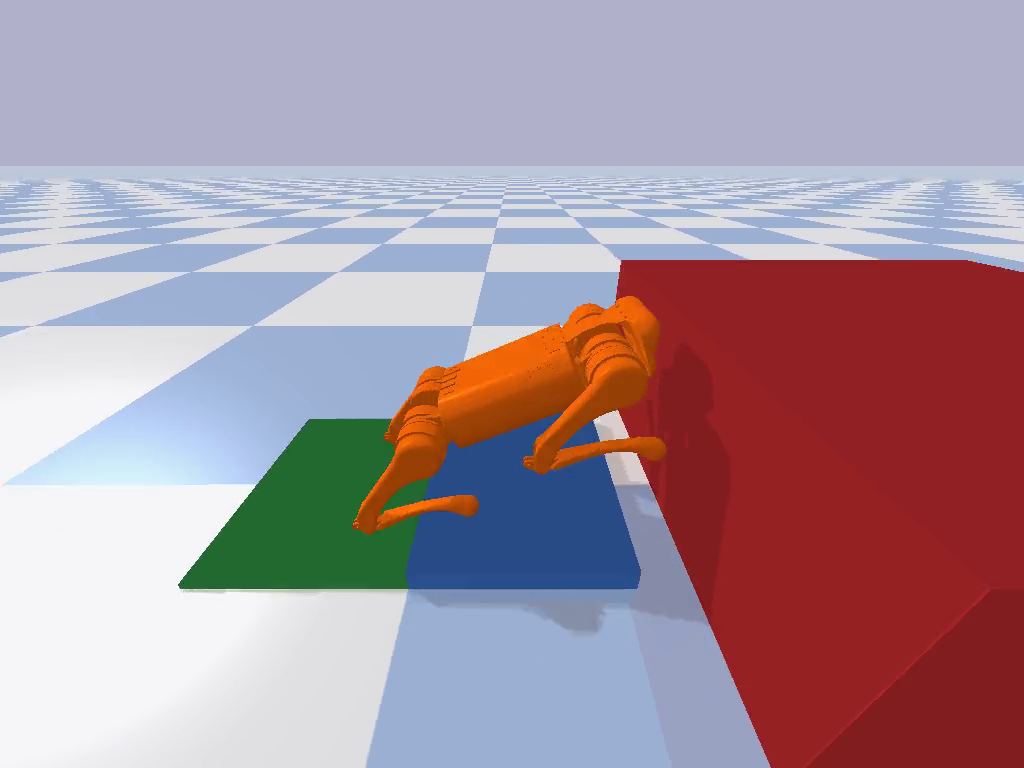}\\
      \caption{Motion snapshots of a jump of distance $0.7\ m$ and height $0.4\ m$. The front feet have a $0.05\ m$ block beneath them, and the rear feet have a $0.01\ m$ block beneath them. \textbf{Top:} The learned policy successfully outputs trajectory offsets to jump onto the platform. \textbf{Bottom:} The feedforward controller results in overpitch and overjumps vertically, falling short of the platform. }
      \label{fig:snapshots}
      \vspace{-0.6em}
\end{figure*}

\begin{figure*}[th]
      \centering
      \includegraphics[width=\snapshotMargin]{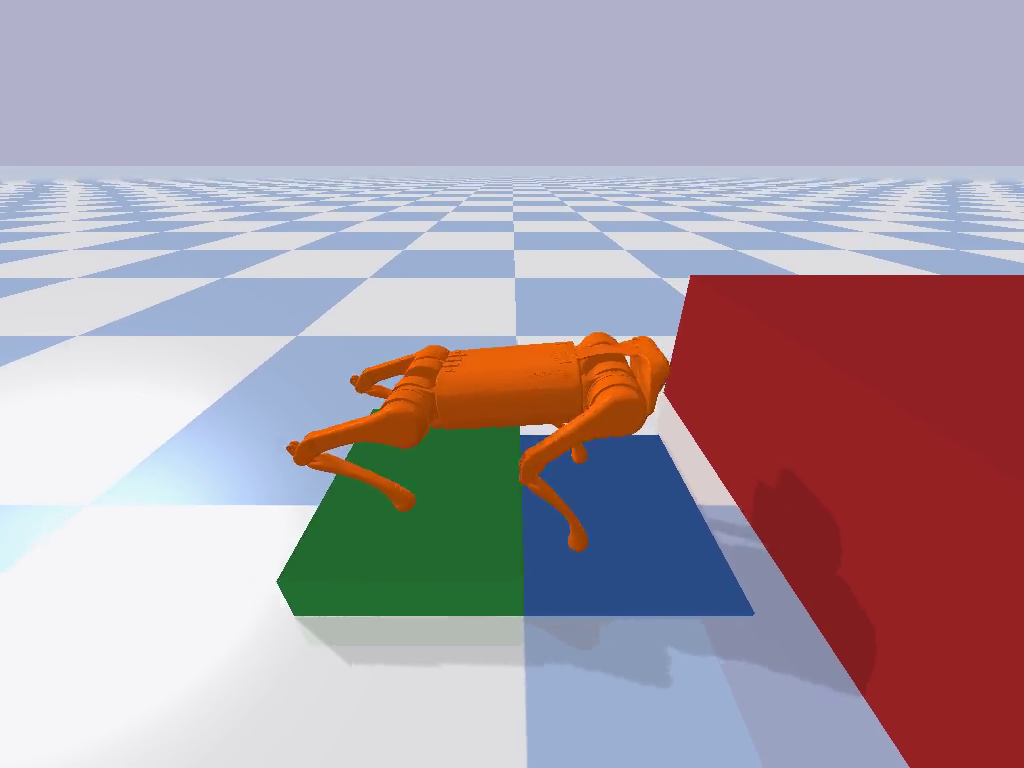}\includegraphics[width=\snapshotMargin]{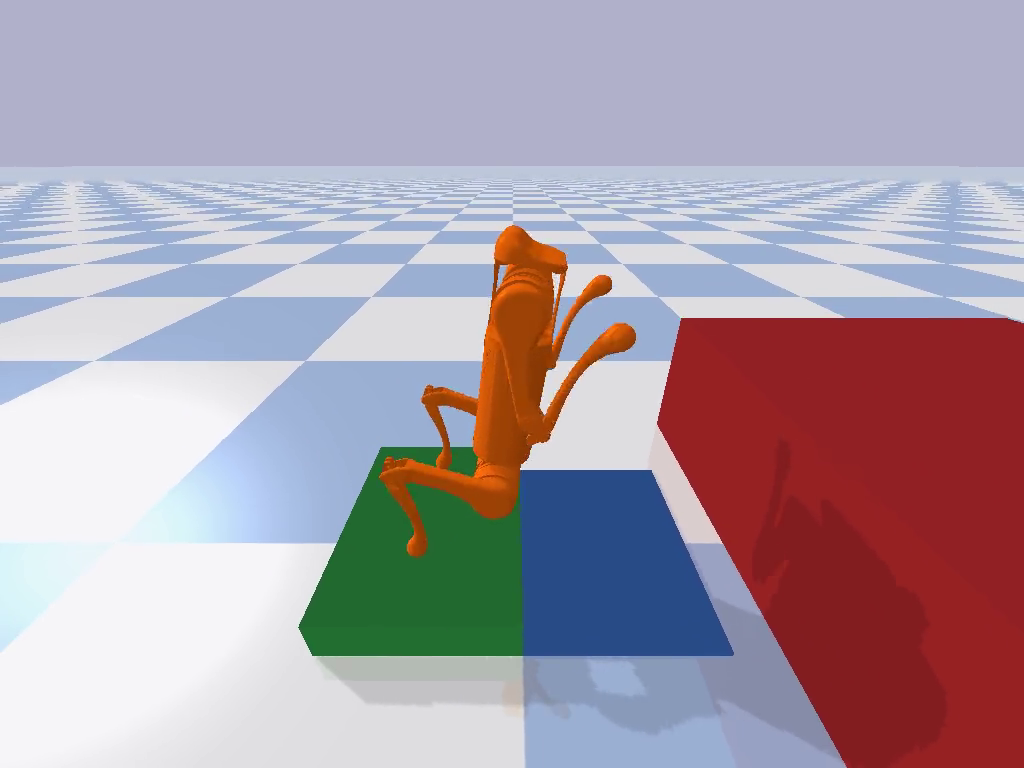}\includegraphics[width=\snapshotMargin]{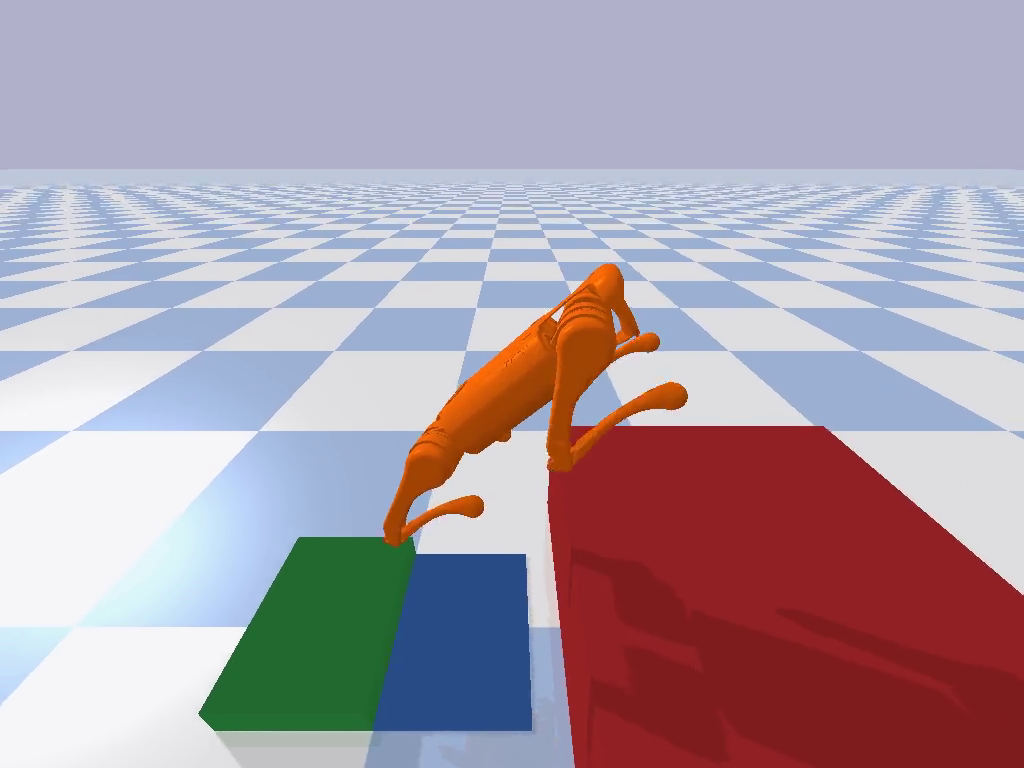}\includegraphics[width=\snapshotMargin]{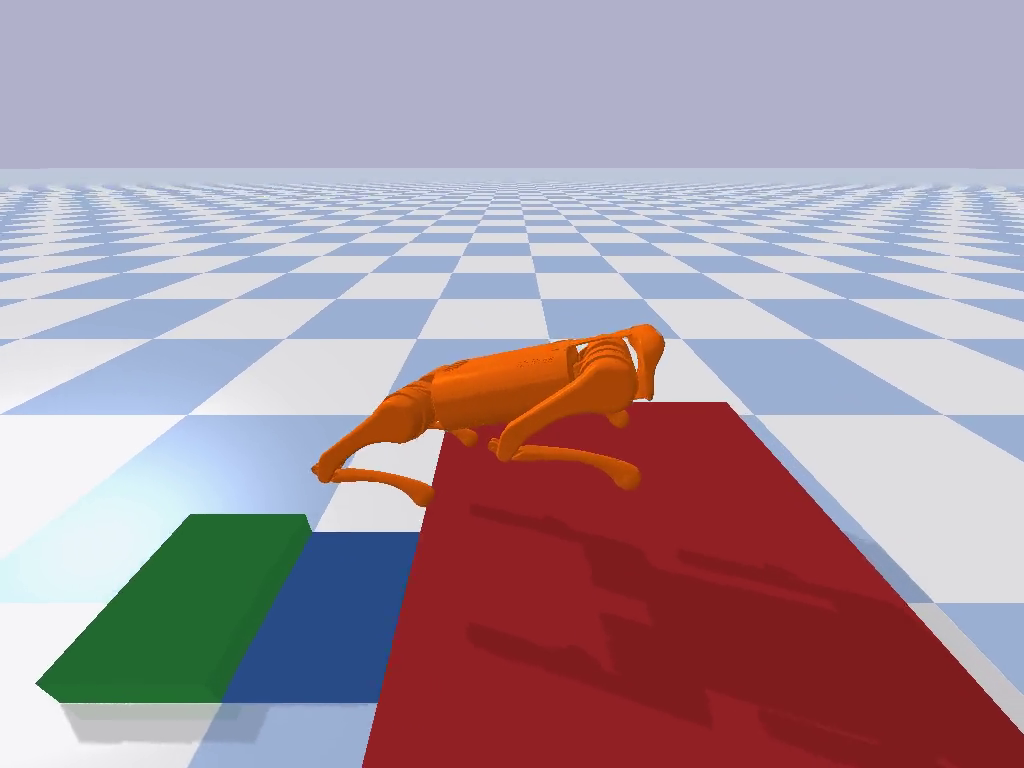}\includegraphics[width=\snapshotMargin]{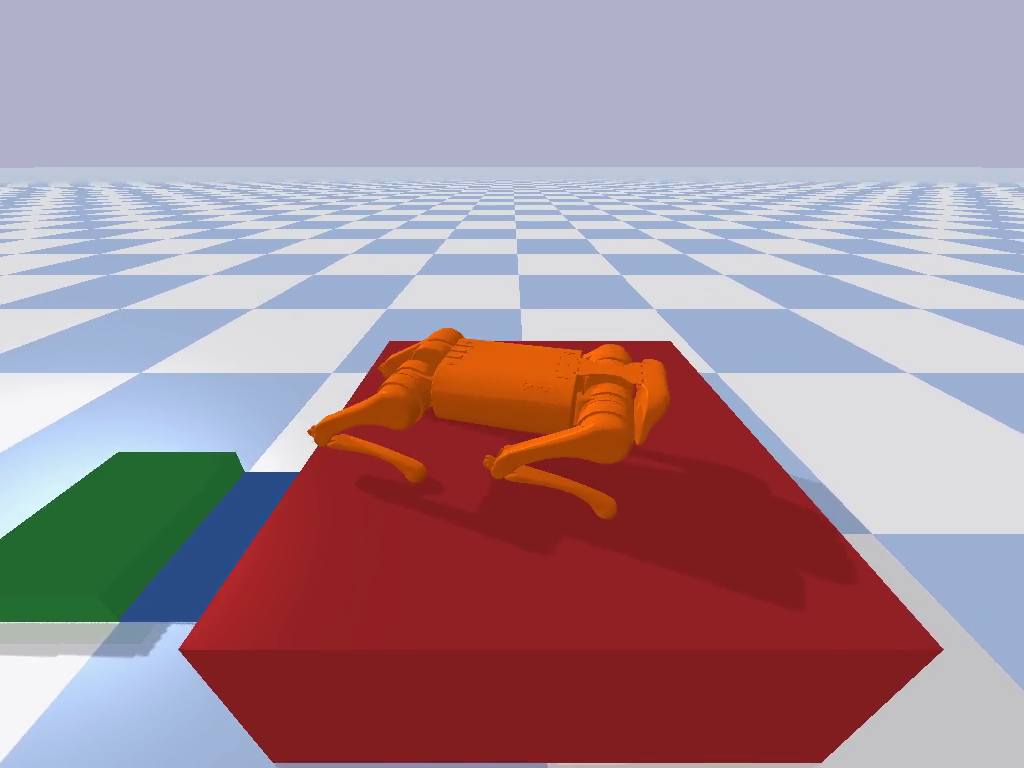}\includegraphics[width=\snapshotMargin]{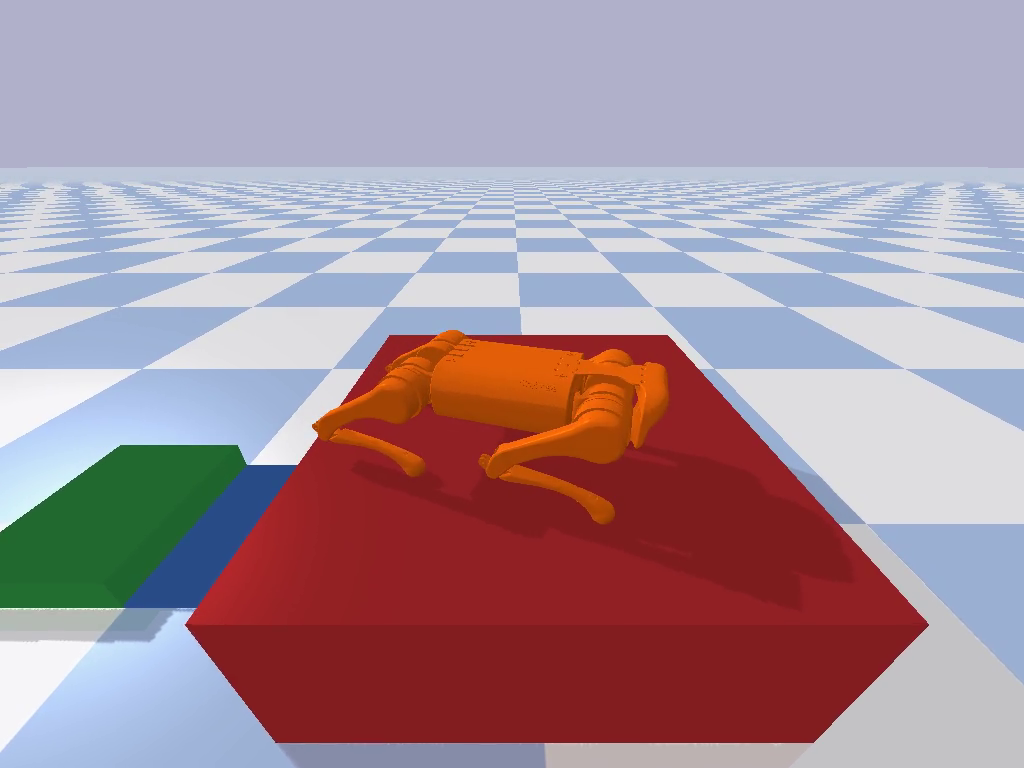}\\
      \vspace{0.1em}
      \includegraphics[width=\snapshotMargin]{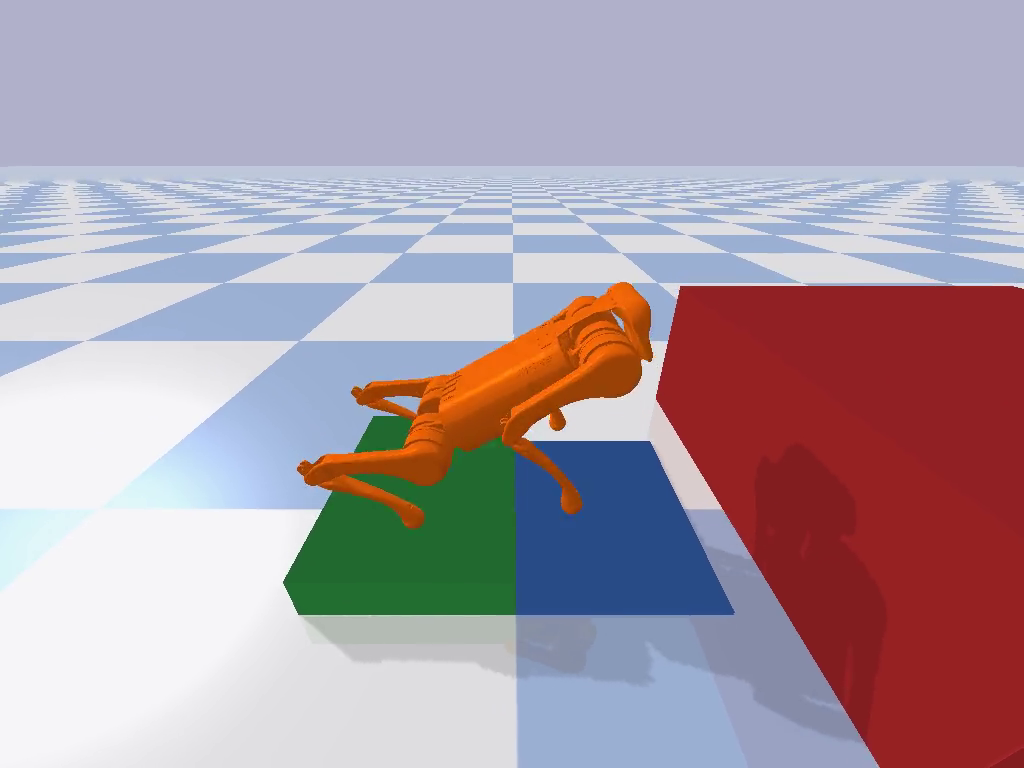}\includegraphics[width=\snapshotMargin]{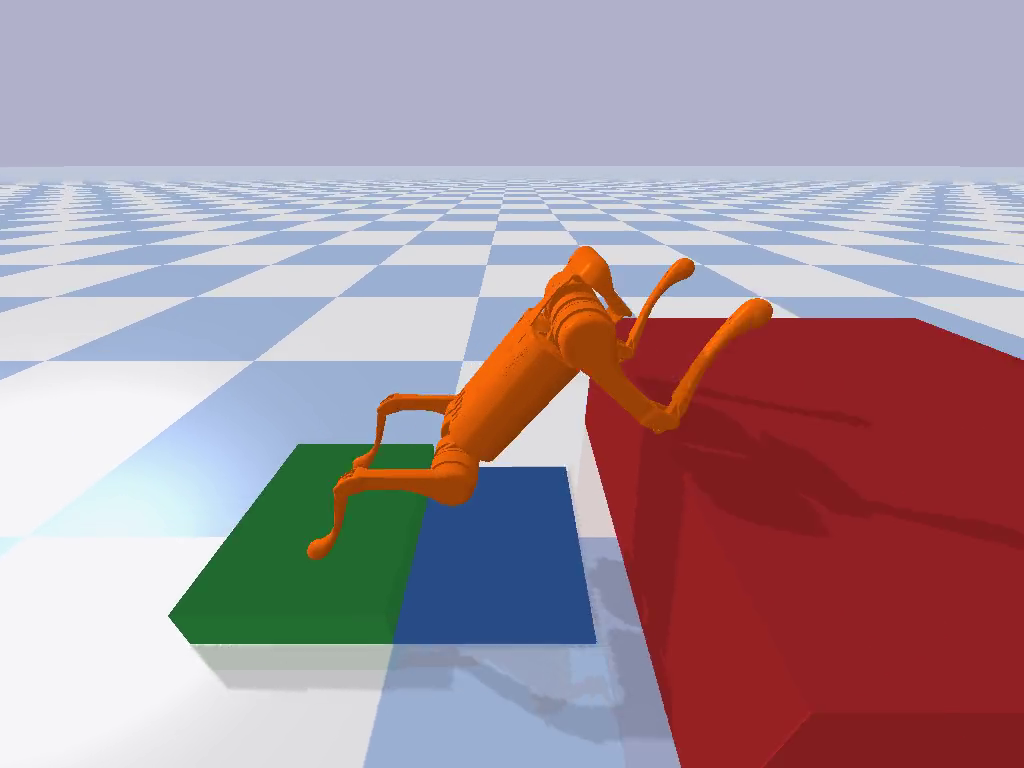}\includegraphics[width=\snapshotMargin]{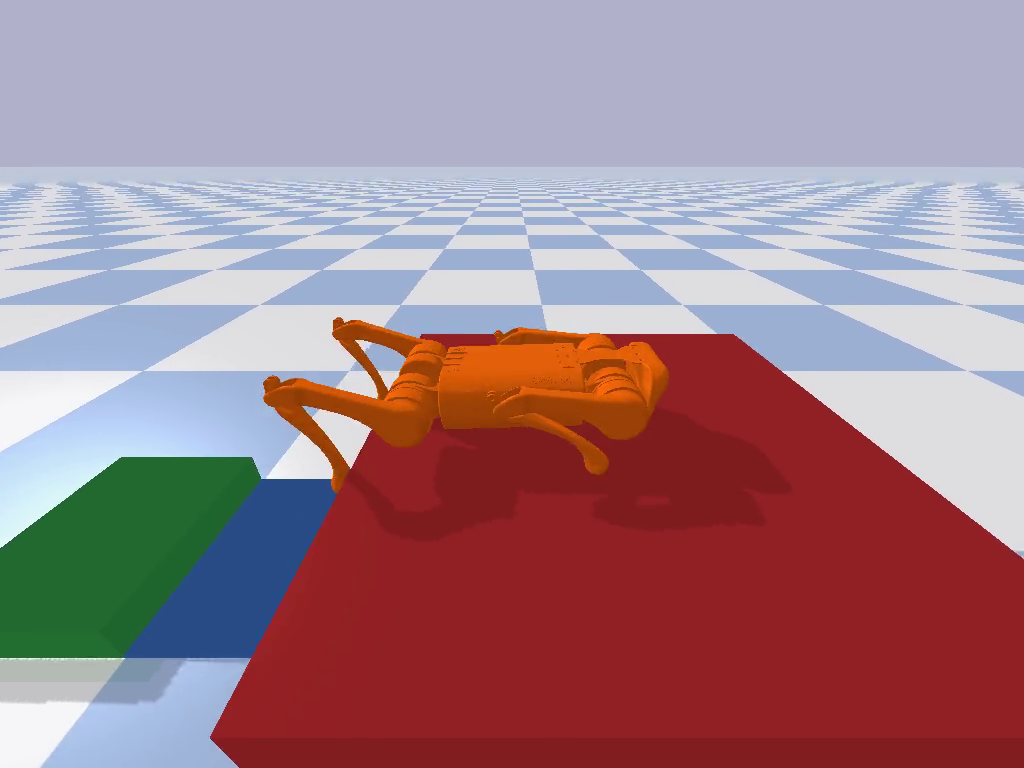}\includegraphics[width=\snapshotMargin]{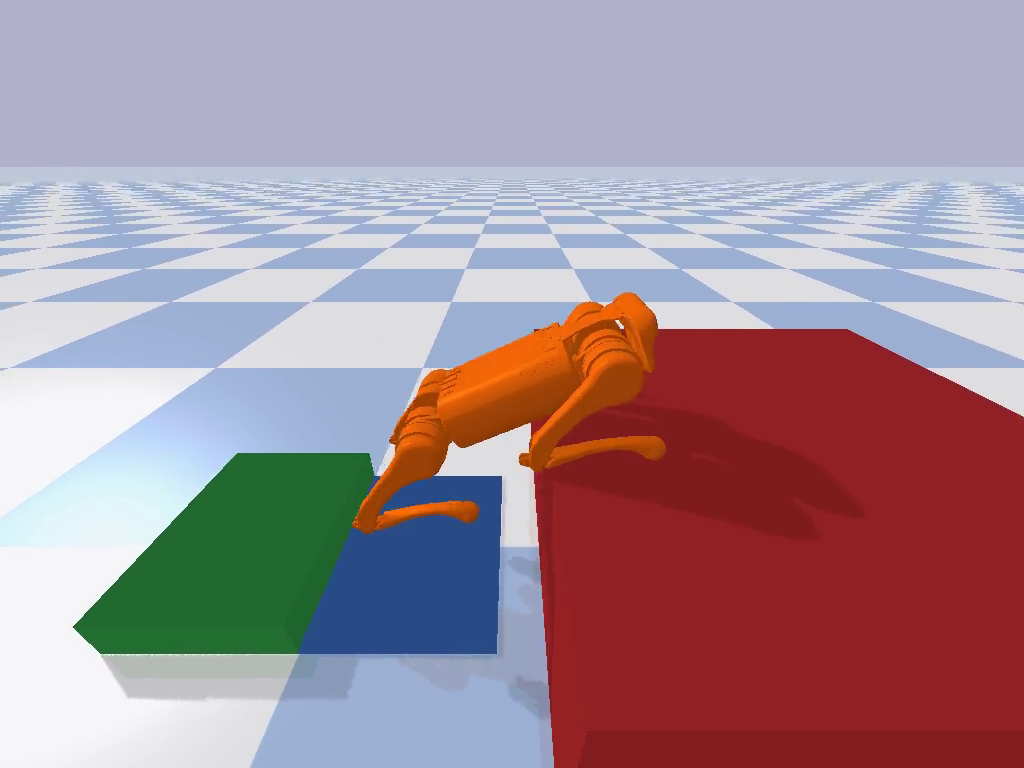}\includegraphics[width=\snapshotMargin]{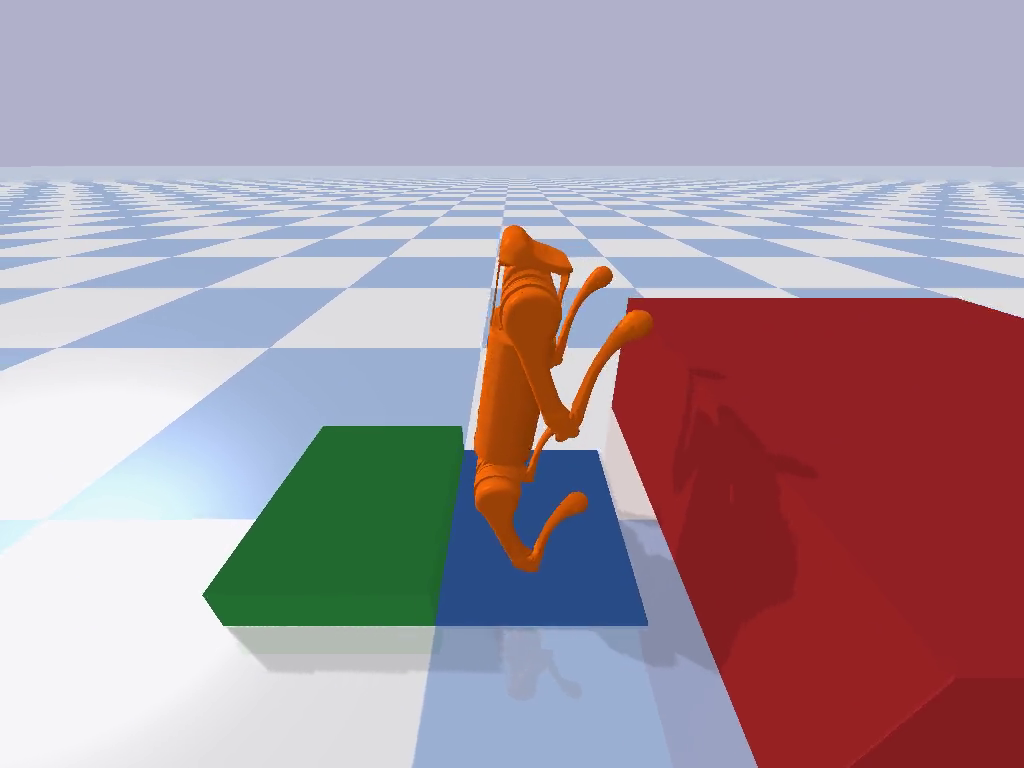}\includegraphics[width=\snapshotMargin]{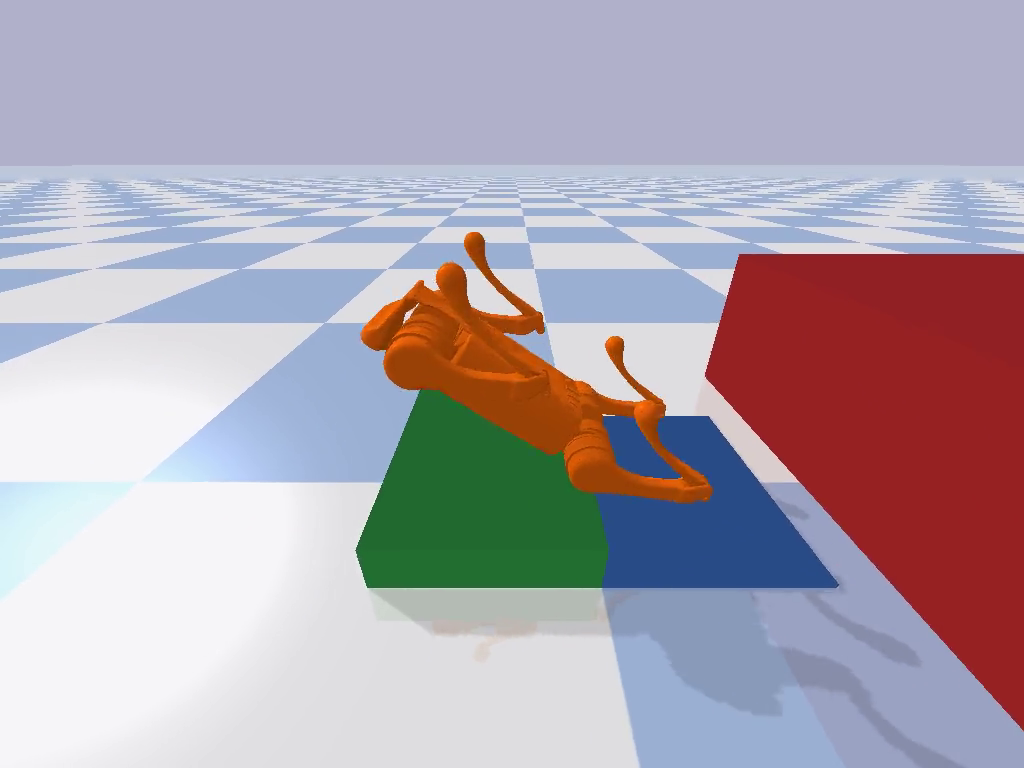}\\
      \caption{Motion snapshots of a jump of distance $0.7\ m$ and height $0.4\ m$. The front feet have a $0.01\ m$ block beneath them, and the rear feet have a $0.1\ m$ block beneath them. \textbf{Top:} The learned policy successfully outputs trajectory offsets to jump onto the platform. \textbf{Bottom:} The feedforward controller results in underpitch and overjumps horizontally, making the rear legs catch on the edge of the platform, resulting in falling off. }
      \label{fig:snapshots_rear}
      \vspace{-1.2em}
\end{figure*}
\begin{figure}[t]
    \centering
    \includegraphics[width=\linewidth]{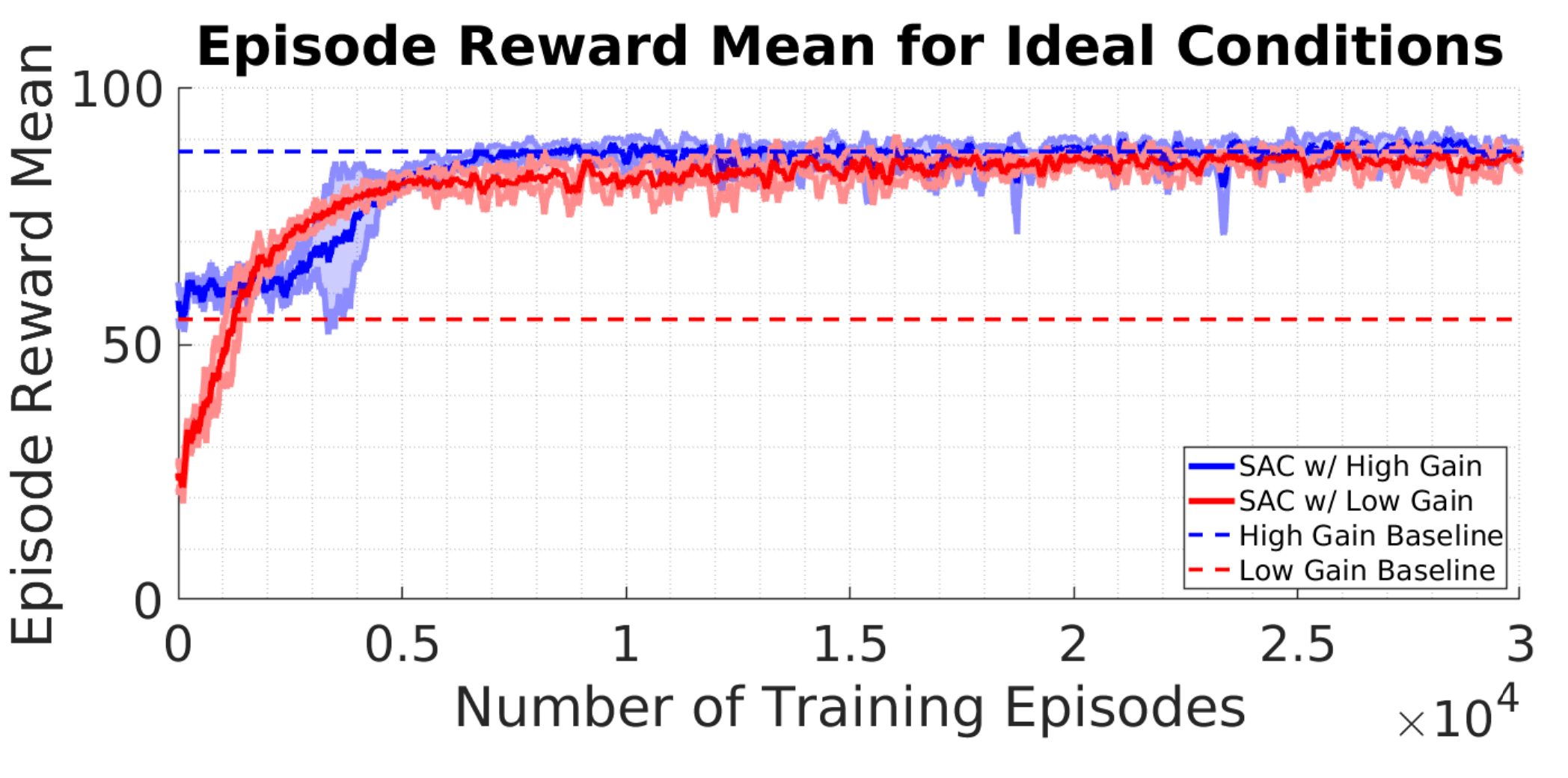} \\
    \caption{Episode reward mean while training under ideal conditions. The baseline feedforward controllers' performance are shown as dotted lines. Our framework is able to track the trajectories accurately for either set of joint gains studied.}
    \label{fig:ideal_conditions}
\end{figure}
\begin{figure}[t]
      \centering
      \includegraphics[width=\linewidth]{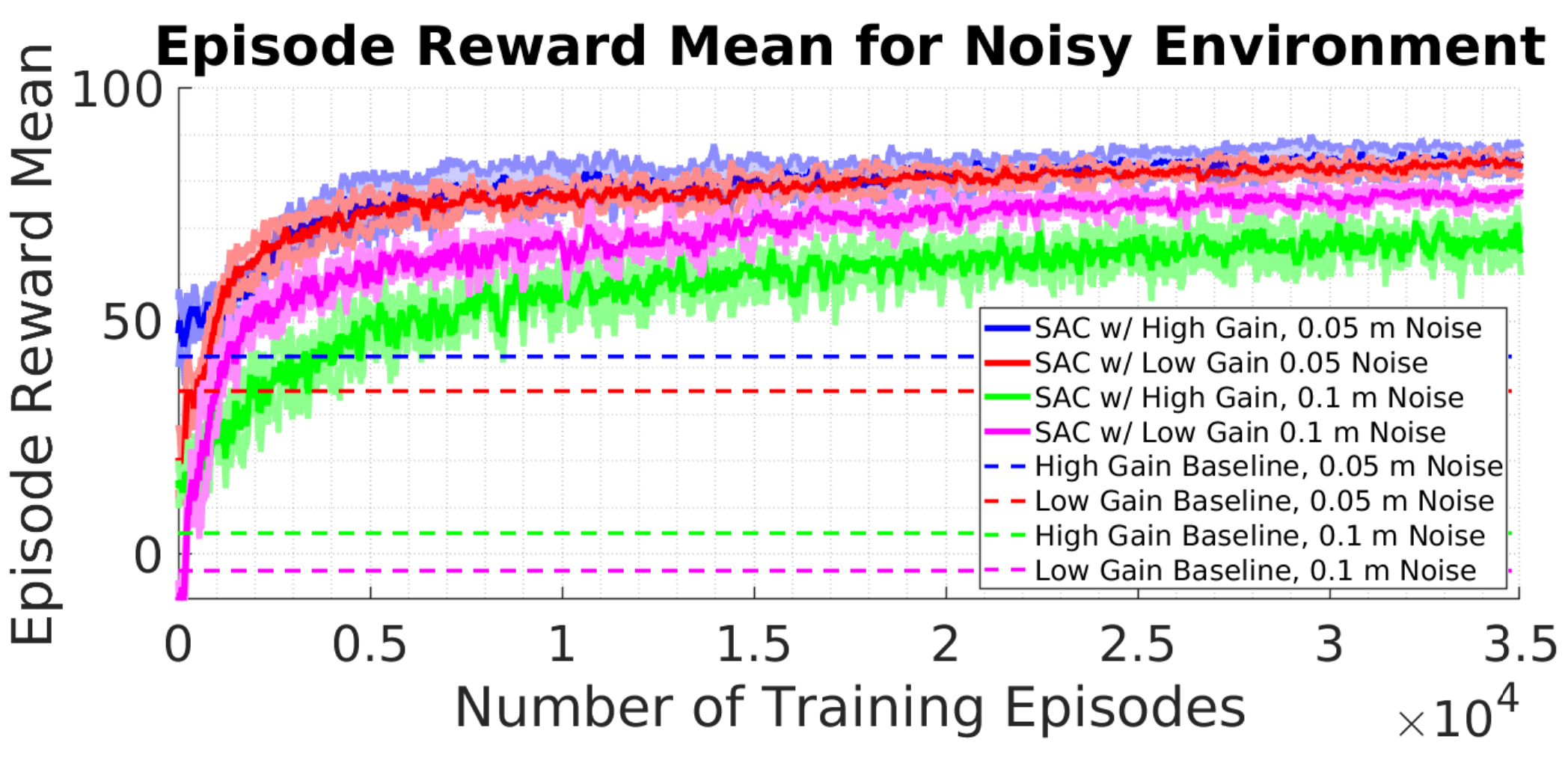} \\
      \caption{Episode reward mean while training with noisy environment conditions: either up to $0.05\ m$ or $0.1\ m$ blocks under each foot, and base mass/inertia vary by up to 5\% of their nominal values. While the feedforward controllers' performance is extremely poor, our method is able to learn to jump accurately through significantly noisy environment conditions. }
      \label{fig:noisy_conditions}
\end{figure}

\begin{figure*}[!t]
      \centering
      \includegraphics[width=\linewidth]{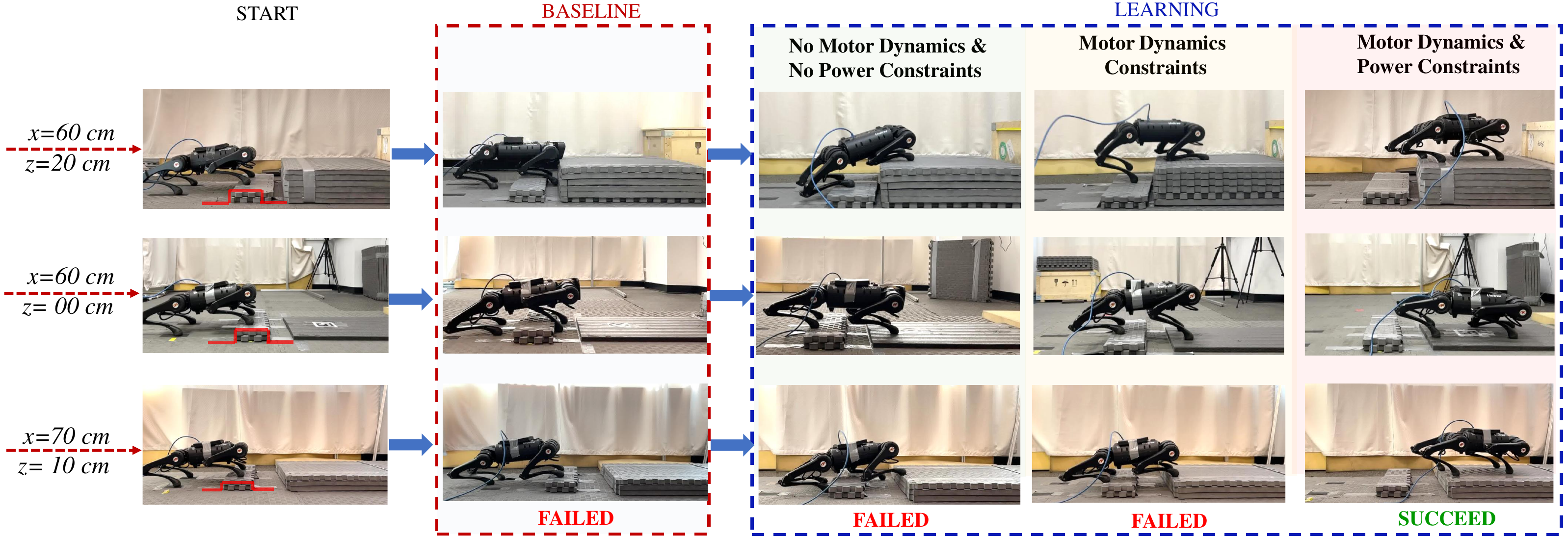} \\
      \caption{\textbf{Jumping different target heights and distances under unknown disturbances}. The figures show baseline experiments (feedforward controller) and sim-to-real transfers for different subsets of motor dynamics and power constraints integration during the learning process. For each jumping target, the robot starts with the same initial configuration, and an unknown disturbance of a $6 cm$ block (red line) is placed under the front feet. We use the same controller gains for all experiments: $K_{p, joint}=300, K_{d,joint}=3$.  Experiment video link: \insertYoutubeLinkHardware.
      }
      \label{fig:all_exp}
      \vspace{-0.5em}
\end{figure*}

In this section, we discuss results from using our method to achieve robust jumping. Example snapshots of the jumping task are shown in Figures~\ref{fig:intro_jump},~\ref{fig:snapshots},~\ref{fig:snapshots_rear}, 
\ref{fig:all_exp},
and the reader is encouraged to watch the supplementary videos\footnote{Simulation video: \insertYoutubeLinkSim}\textsuperscript{,}\footnote{Experiment video: \insertYoutubeLinkHardware} for clearer visualizations. In particular, we show the results of zero-shot sim-to-real transfers of the learned trajectory-offset policies from PyBullet to the Unitree A1 hardware. 

For our experiments, we are specifically interested in the following questions: 
\begin{itemize}
    \item [1.] How does choice of joint gain affect tracking performance? 
    \item [2.] Can we improve upon tracking performance in ideal conditions? 
    \item [3.] How does (magnitude of) noise affect the agent's ability to learn?
    \item [4.] What is the importance of integrating both motor dynamics constraints and power constraints into the learning environment? 
\end{itemize}
We consider two different sets of joint gains, which we name ``high'' $(\bm{K}_{p,joint}=300\bm{I}_3,\ \bm{K}_{d,joint}=3\bm{I}_3)$ and ``low'' $(\bm{K}_{p,joint}=100\bm{I}_3,\ \bm{K}_{d,joint}=2\bm{I}_3)$ gains. Oftentimes these gains must be tuned by hand, and may also need to be adapted for different trajectories. Thus, our goal with learning with different gains is that it may give some insight on if we can indirectly tune these all at once for multiple trajectories by selecting trajectory offsets, rather than manual human trial and error. We set the Cartesian gains as $\bm{K}_{p}=500\bm{I}_3,\ \bm{K}_{d}=10\bm{I}_3$.

\subsection{Simulation Results}

Figure~\ref{fig:ideal_conditions} shows training results for learning to offset the trajectories under ideal conditions. With the default baseline ``high'' joint gain jumping controller, the tracking is already very accurate, getting close to $w=100$ rewards. On the other hand, the baseline ``low'' gain controller does not perform as well, primarily due to errors in pitch when at the end of the trajectory, as well as falling short in distance. However, through our framework, we are able to accurately track the desired jumping motions using either set of gains, though the ``high'' gains still result in slightly better performance. This shows that our framework is general enough to learn to improve several jumping behaviors without the need to explicitly tune gains on a per-motion basis, as may often be needed in general. 

Figure~\ref{fig:noisy_conditions} shows training results for learning to offset the trajectories under the noisy conditions described in~\ref{sec:training_details}. We train under two sets of noisy environment conditions: with either up to $0.05\ m$ height noise, or up to $0.1\ m$ height noise under each foot, in addition to the variability in body mass and inertia. The baseline controllers for either set of gains are not able to accurately track the desired motions, predominantly due to over/under pitching during the single contact rear back phase. This becomes especially apparent as we increase the environment noise to $0.1\ m$, where under our reward scheme, the feedforward controller averages approximately 0 reward across 100 random trials, for either set of joint gains, corresponding to extremely poor performance where the robot is not even close to the goal location. 

The bottom row of Figure~\ref{fig:snapshots} shows the over-pitching behavior of the baseline controller when the front legs start higher than the rear ones, during one of the more difficult jumps in terms of height and distance. This results in jumping vertically and not coming close to landing on the platform. 

The bottom of Figure~\ref{fig:snapshots_rear} shows the opposite result (under-pitching) when the rear feet start at a higher $z$ height than that of the front feet. In this case, the baseline feedforward controller does not pitch enough before take off, leading to a more horizontal jump that crashes horizontally into the platform. For both of these scenarios, our learned controller is able to successfully jump onto the platform, as can be seen in the top rows of Figures~\ref{fig:snapshots} and~\ref{fig:snapshots_rear}.  

A noteworthy observation is that while the ``low'' gain baseline performance (as well as when training with our method) is not as good as the ``high'' gain controller for ideal conditions, as the noise increases significantly, we see that the agent is able to exploit the lower gain joint controller to outperform the policy using the high gain controller, as can be seen in Fig.~\ref{fig:noisy_conditions}.

These results show that through our method, using either set of gains, we are able to learn offsets to significantly improve jumping performance under noisy environmental conditions, close to as well as under ideal conditions.

\subsection{Experimental Verification}
\label{sec:experiments}

We validate the effectiveness of our proposed learning framework, which incorporates motor dynamics and power constraints, in enabling robust jumping on the Unitree A1 robot hardware. We conduct various experiments with different jumping targets of $(x,z)\in \lbrace (60,20),(60,0),(70,10) \rbrace (cm)$, and different block disturbances  $\lbrace 3, 6 \rbrace (cm)$ introduced under the robot's feet. 
We focus our discussion on the $6cm$ disturbance, as illustrated in Fig. \ref{fig:all_exp}. This disturbance amounts to $33 \%$ of the robot's initial height and is not explicitly known by the agent. 

In the baseline experiments, we only use the joint PD and Cartesian PD controller to track the joint and foot trajectory references from the full-body trajectory optimization, as described in Section \ref{sec:learning_TO}. Since there is no feedback controller to compensate for the uneven terrain disturbance, robot trajectory errors propagate during the jumping process, resulting in overpitching before take-off and failed jumps (Fig. \ref{fig:all_exp}).

In contrast to the baseline, our learned jumping policy is run as a real-time feedback controller to compensate jumping trajectory errors. In order to verify the effectiveness of integrating motor dynamics and power constraints into the learning environment, we compare the sim-to-real transfer performance of training controllers with the following subset of constraints: 
\begin{itemize}
    \item [1.] No motor dynamics nor power constraints.
    \item [2.] Only motor dynamics constraints.
    \item [3.] Both motor dynamics and power constraints.
\end{itemize}

\begin{figure}[t]
	\centering
 \vspace{-1.5em}
  \includegraphics[width=1\linewidth,trim={1cm 0cm 1cm 0cm},clip]{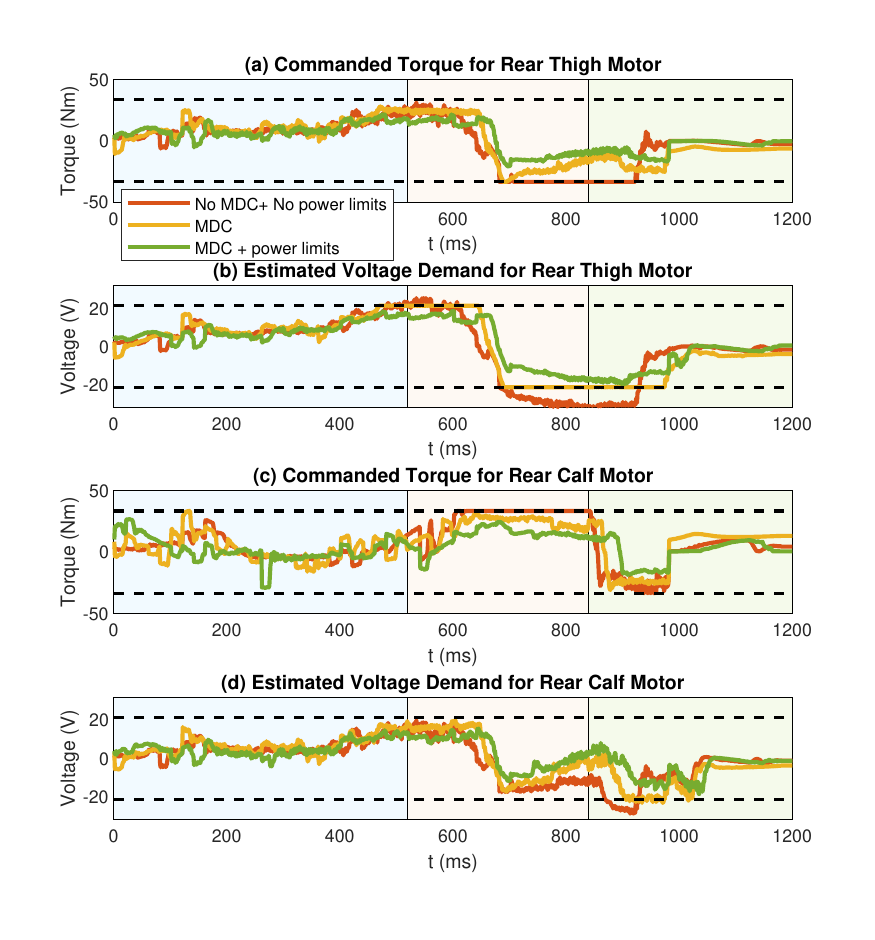}
	\\
    \vspace{-2em}
	\caption{\textbf{Experiments for jumping on box $\bm{(x,z)=(60,20) cm}$}. Estimated voltage demands and total torque commands for the Rear Right leg thigh and calf motors. For this motion, the rear legs typically require more torques than front legs, so it is sufficient to consider the torque and voltage profiles for the rear legs. The maximum motor torques and battery voltage are specified by dotted black lines. The single-leg contact and flight phases start at approximately $520~ms$ and $840~ms$, which are specified as vertical black lines.
    } 
    \vspace{-0.8em}
	\label{fig:comparison_exp_mdc_plots}
\end{figure}

\begin{figure}[t]
	\centering
  \includegraphics[width=1\linewidth,trim={0.6cm 0cm 1cm 0cm},clip]{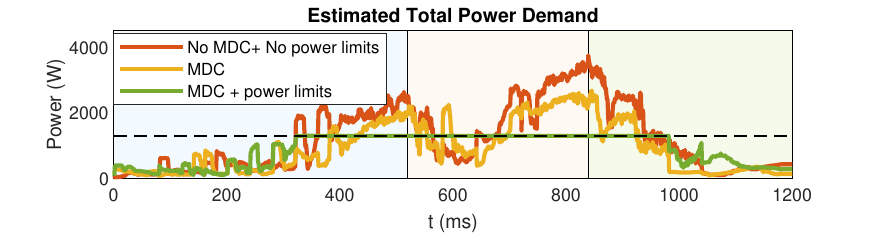}\\
	\caption{
Estimated total power requirement to operate all motors for jumping onto a {$(x,z)=(60,20) cm$} box, associated with Fig. \ref{fig:comparison_exp_mdc_plots}. The dotted black line represents the maximum power of the onboard battery supply.
 } 
    \vspace{-0.8em}
	\label{fig:comparison_exp_power_plots}
\end{figure}

\textit{Case 1 - Learning with No Constraints}: In the first learning experiment for each target, we only consider a na\"ive implementation of torque limits that is widely utilized for learning locomotion (e.g., \cite{tan2018minitaur,peng2020laikagoimitation, bellegarda2021robust,bellegarda2022cpgrl}), in which only a saturation function is applied for the final torque command.
As can be seen in Fig. \ref{fig:all_exp}, the robot has learned to compensate for an unknown noise of a $6cm$ block, thereby jumping farther than the baseline cases. However, jumping onto the $(x,z)=(60,20) cm$ box, for example, requires high voltage of up to $30V$ for the motors of the rear right leg at the time of
taking off, as illustrated in Fig. \ref{fig:comparison_exp_mdc_plots}b and \ref{fig:comparison_exp_mdc_plots}d (roughly at $840~ms$). It also demands a significant total power supply of approximately $3750W$ (Fig. \ref{fig:comparison_exp_power_plots}). These requirements exceed the battery capability
of $V_{max} = 21.5V$, $P_{max} = 1290W$ (i.e.~violates both motor dynamics and battery power limits).
These violations cause poor tracking performance and result in the robot falling short of the target, as can be seen in Fig. \ref{fig:all_exp}. 

\textit{Case 2 - Learning with Only MDC}: For the second learning experiment, our policy learns to output actions which are then constrained by the torque-speed relationship throughout the whole jump, as described in Equation (\ref{eq:MDC_voltage}). Therefore, the voltages for both thigh and calf motors are always within the limits (Fig. \ref{fig:comparison_exp_mdc_plots}b and \ref{fig:comparison_exp_mdc_plots}d). However, the robot still fails to reach the target because this aggressive motion did not consider the power limits. In particular, this jumping motion requires up to $2500 W$ when taking off, which is nearly double the maximum battery power (Fig. \ref{fig:comparison_exp_power_plots}).

\textit{Case 3 - Learning with MDC and Power Constraints}: The third learning experiment demonstrates the importance of considering both Motor Dynamic Constraints and Power Constraints in order to realize successful sim-to-real transfers for highly aggressive jumping maneuvers. The proposed integration of both motor dynamics and power limits ensure (i) the voltage demanded for operating the motors can be supplied by the battery and (ii) the required total power for all motors satisfies the on-board power supply. Both of these limits can be verified to not be violated in Figures \ref{fig:comparison_exp_mdc_plots}b,  \ref{fig:comparison_exp_mdc_plots}d, and \ref{fig:comparison_exp_power_plots} for the example target jump of $(x,z)=(60,20) cm$. Additionally, our method enables the robot to successfully reach various jumping targets while ensuring robustness against unknown and large disturbances (e.g. $6cm$ block), as illustrated in Fig. \ref{fig:all_exp}.

\begin{figure}[t]
	\centering
  \includegraphics[width=1\linewidth,trim={0.6cm 0cm 1cm 0cm},clip]{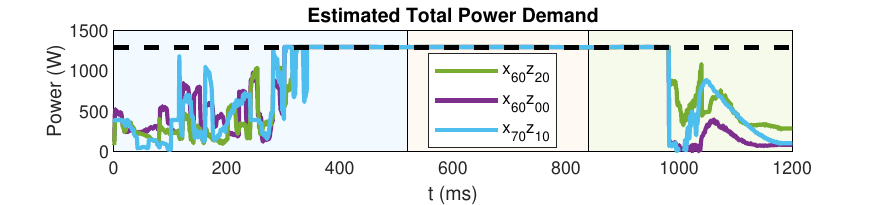}\\
	\caption{
 \textbf{Experiments for different jumping targets.}
 The estimated total power demand for jumping to different targets with policies trained with both motor dynamics and power constraints, corresponding to Fig. \ref{fig:all_exp}. The maximum power of the battery ($P_{max}\approx 1290 W$) is specified as the dotted black line.
 } 
    \vspace{-0.8em}
	\label{fig:comparison_power_targets}
\end{figure}

A noteworthy observation from Fig. \ref{fig:comparison_power_targets} is that all jumping motions rapidly reach the battery power limits, even for jumping forward without a desired height goal $(x,z)=(60,0) cm$. Our learning framework, which integrates motor dynamics and power limits, provides a practical solution to achieve various jumping targets despite the limited power capacity of the onboard battery.

\section{Conclusion}
\label{sec:conclusion}

In this work, we have proposed a method to improve jumping performance of optimal trajectories with deep reinforcement learning. Instead of learning from scratch, we learn to modify and augment the existing trajectories in Cartesian space, which proved to be robust to significantly varying environmental conditions. In addition to robustness to environmental perturbations, we showed robustness to different joint gains, and further benefits include avoiding re-running potentially expensive optimization routines at run time for changes to the initial robot state, uncertainty in the system dynamics, and uncertainty in the environment such as varying uneven terrain and coefficients of friction. In order to realize highly aggressive jumps on hardware, we proposed and integrated motor dynamics and power limits as key components of the learning environment, enabling effective sim-to-real transfers without any further tuning. Our results demonstrate full exploitation of the available hardware power and motor limits to jump twice the body length in distance while subject to uneven terrain noise of 33\% of the nominal standing height. 

\section*{Acknowledgments}
The authors would like to thank Zhuochen Liu and Yiyu Chen from the Dynamic Robotics and Control Laboratory for the insightful discussion on the experimental setup.

\bibliographystyle{IEEEtran}
\bibliography{refs} 

\end{document}